\documentclass[runningheads]{llncs}
\usepackage{graphicx}
\usepackage{amsmath,amssymb} 
\usepackage{color}
\usepackage[width=122mm,left=12mm,paperwidth=146mm,height=193mm,top=12mm,paperheight=217mm]{geometry}

\usepackage{algorithm}
\usepackage{multirow}
\usepackage{algorithmic}
\usepackage{subfigure}
\usepackage{enumerate}

\newcommand\ie{\emph{i.e.}} 
 
\newcommand\etal{\emph{et al.}}

\newcommand{\rank}{\mbox{\rm rank}}

\newcommand{\SKIP}[1]{} 

\newcommand{\mbegin} {\left [ \begin{array}}

\newcommand{\mend}   {\end{array} \right ]}

\newcommand{\detbegin} {\left | \begin{array}}
\newcommand{\detend}   {\end{array} \right |}

\newcommand{\vbegin} {\left ( \begin{array}{c}}

\newcommand{\vend} {\end{array}\right )}

\def\squareforqed{\hbox{\rlap{$\sqcap$}$\sqcup$}}

\def\qed{\ifmmode\squareforqed\else{\unskip\nobreak\hfil
	\penalty50\hskip1em\null\nobreak\hfil\squareforqed
	\parfillskip=0pt\finalhyphendemerits=0\endgraf}\fi}

\newcommand{\showeqnlabel}{
	\hbox to 0pt{\quad\quad\relax\fbox{\scriptsize\rm\eqnlblx}%
	\gdef\eqnlblx{xxxx}}} \newcommand{\eqnlblx}{}

\def\@eqnnum{\rm (\theequation)\showeqnlabel}

\newcommand{\nofig}[1]{\centerline{\bf Figure here}}

\def\mat#1{\mathchoice{\mbox{\bf$\displaystyle\tt#1$}}
	{\mbox{\bf$\textstyle\tt#1$}}
	{\mbox{\bf$\scriptstyle\tt#1$}}
	{\mbox{\bf$\scriptscriptstyle\tt#1$}}}
\def\m#1{\protect\mat #1}

\begin{document}

\pagestyle{headings}
\mainmatter
\def\ECCV16SubNumber{1858}  

\title{Multi-body Non-rigid Structure-from-Motion} 


\authorrunning{Suryansh Kumar, Yuchao Dai, and Hongdong Li}

\author{Suryansh Kumar$^{1}$, Yuchao Dai$^{1}$, and Hongdong Li$^{1, 2}$}
\institute{$^1$Research School of Engineering, the Australian National University \\ $^2$ Australian Centre for Robotic Vision. }

\maketitle

\begin{abstract}
Conventional structure-from-motion (SFM) research is primarily concerned with the 3D reconstruction of a single, rigidly moving object seen by a static camera, or a static and rigid scene observed by a moving camera --in both cases there are only one relative rigid motion involved.  Recent progress have extended SFM to the areas of {\em multi-body SFM} (where there are {\em multiple rigid} relative motions in the scene), as well as {\em non-rigid SFM} (where there is a single non-rigid, deformable object or scene).  Along this line of thinking, there is apparently a missing gap of ``multi-body non-rigid SFM", in which the task would be to jointly reconstruct and segment multiple 3D structures of the multiple, non-rigid objects or deformable scenes from images. Such a multi-body non-rigid scenario is common in reality (e.g. two persons shaking hands, multi-person social event), and how to solve it represents a natural {\em next-step} in SFM research. By leveraging recent results of subspace clustering, this paper proposes, for the first time, an effective framework for multi-body NRSFM, which simultaneously reconstructs and segments each 3D trajectory into their respective low-dimensional subspace. Under our formulation, 3D trajectories for each non-rigid structure can be well approximated with a sparse affine combination of other 3D trajectories from the same structure (self-expressiveness). We solve the resultant optimization with the alternating direction method of multipliers (ADMM). We demonstrate the efficacy of the proposed framework through extensive experiments on both synthetic and real data sequences. Our method clearly outperforms other alternative methods, such as first clustering the 2D feature tracks to groups and then doing non-rigid reconstruction in each group or first conducting 3D reconstruction by using single subspace assumption and then clustering the 3D trajectories into groups.
\SKIP{Although the developed framework is non-convex, under convex relaxation, the method succeeds in recovering desired non-rigid 3D reconstruction and its sparse subspace representation.} 
\end{abstract}

\begin{figure}
\centering
\includegraphics[width=1.0\textwidth] {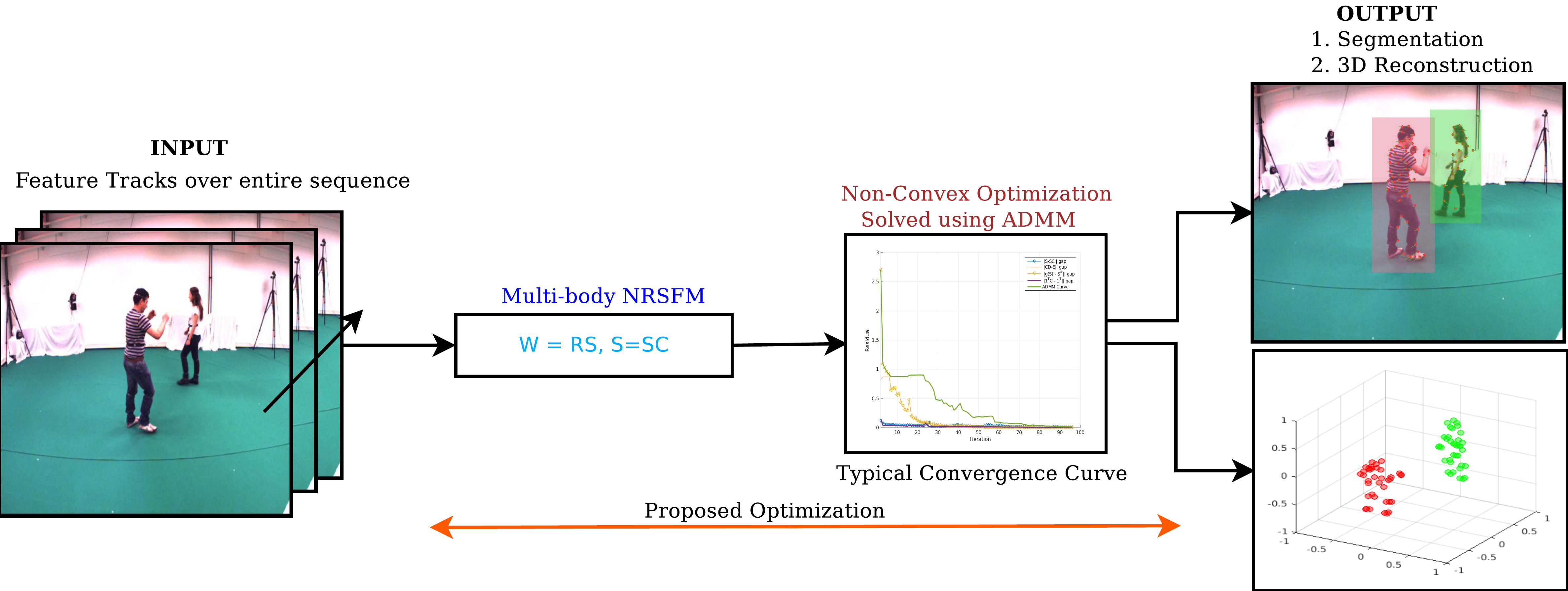}~~~
\caption{Our framework takes 2D feature tracks of objects undergoing non-rigid deformation over entire image sequence as input and outputs its 3D reconstruction and segmentation. We represent each non-rigid object motion spans as lying in an affine subspace. Although, such assumption leads to non-convex optimization problem but we propose an efficient procedure to solve it via ADMM. Here, two subjects are performing complex non-rigid motion, our framework is able to faithfully segment and reconstruct both of them. The above images are taken from the UMPM dataset \cite{UMPM}.}
\label{fig:flow_diagram}
\noindent\makebox[\linewidth]{\rule{0.83\paperwidth}{0.4pt}}
\end{figure}


\section{Introduction}
Structure-from-Motion targets at recovering 3D structure and camera motion from monocular 2D feature tracks. Conventional SFM primarily concerns with the 3D reconstruction of a single rigidly moving object seen by a static camera, or a static and rigid scene observed by a moving camera --in both cases there are only one relative rigid motion involved. SFM has been extended to the areas of {\em multi-body} SFM, as well as {\em single body} {non-rigid SFM} (see Table.\ref{tab:summary} for a classification of different SFM problems).
Since, NRSFM is central to many computer vision applications such as motion capture, activity recognition, human-computer interaction, and dynamic scene reconstruction, etc. Therefore, it has been extensively researched (e.g. in \cite{Bregler:CVPR-2000}\cite{Xiao-Chai-Kanade:ECCV-2004}\cite{Akhter-Sheikh-Khan-Kanade:Trajectory-Space-2010}\cite{Dai-Li-He:CVPR-2012}\cite{Dense-NRSFM:CVPR-2013}\cite{PND-Mixture-Model:IJCV-2016}) in the past.



\begin{figure}[!htp]
  \begin{center} 
  \subfigure[\label{fig:dance_yoga_2D_1}]{\includegraphics[width=0.24\linewidth]{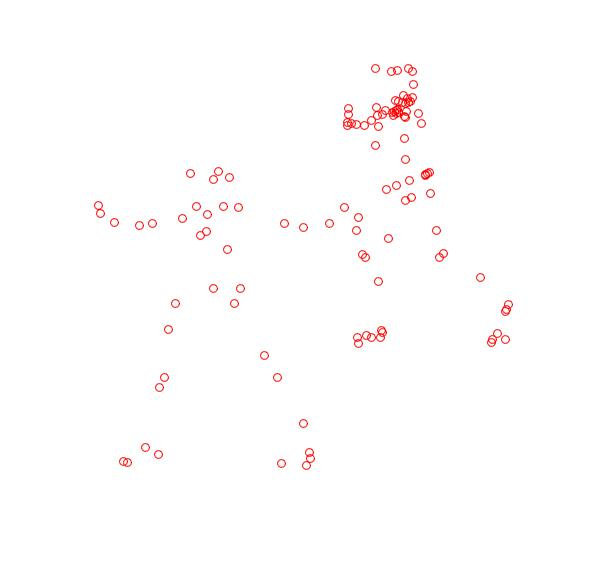}}
  \subfigure[\label{fig:dance_yoga_2D_2}]{\includegraphics[width=0.24\linewidth]{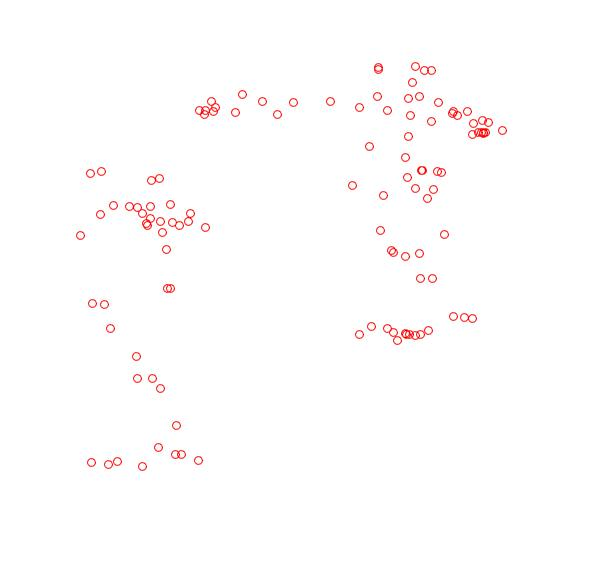}}
  \subfigure[\label{fig:dance_yoga_3D_1}]{\includegraphics[width=0.24\linewidth]{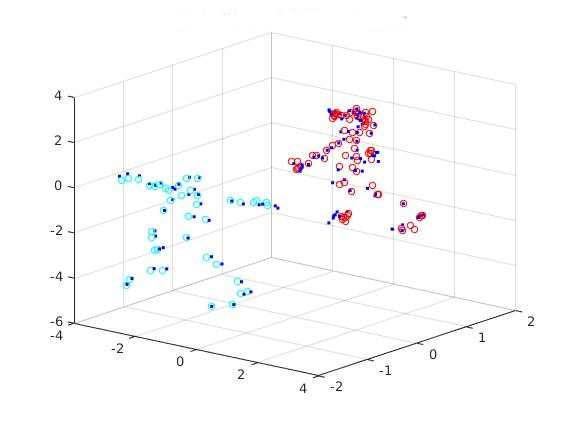}}
  \subfigure[\label{fig:dance_yoga_3D_2}]{\includegraphics[width=0.24\linewidth]{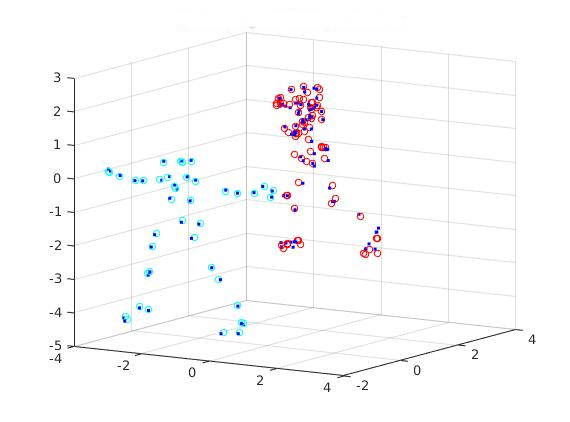}}  
  \end{center}
  \caption{\small Illustration of complex multi-body non-rigid 3D reconstruction, where two persons perform different activities (Dancing and Yoga \cite{Akhter-Sheikh-Khan-Kanade:Trajectory-Space-2010}). Given the multi-frame 2D feature tracks as input, our method simultaneously outputs the 3D non-rigid reconstruction and the segmentation of the track trajectory in 3D. (a) First frame of the 2D track; (b) Second frame of the 2D track; (c) 3D reconstruction and segmentation of (a) using our method, where different colors index shows the corresponding segmentation; (d) Similarly, 3D reconstruction and segmentation of (b). }
  \noindent\makebox[\linewidth]{\rule{0.83\paperwidth}{0.4pt}}
  \label{fig:multiple_nonrigid}
\end{figure}

So far, all existing methods for NRSFM have implicitly assumed that there is only one deformable shape or object in the view. However, the real world scenarios can be much more complicated involving multiple, independently deforming objects in the scene. Multiple nonrigid objects are commonly encountered in our daily lives, for example, in motion capture, multiple persons perform different activities with possible interactions (see Fig.~\ref{fig:multiple_nonrigid} for an example); in human-computer interaction, different users may conduct different gesture commands; and in traffic scene, multiple vehicles and walking pedestrians create {\bf multi-body non-rigid deformations}.  

To handle such multiple non-rigid deformations in 3D reconstruction, a natural idea would be to simply treat the multiple non-rigid deformations as a single (though more complex) non-rigid deformation (with higher order or higher rank), and then apply any state-of-the-art non-rigid structure-from-motion methods such as \cite{Dai-Li-He:CVPR-2012}\cite{Procrustean-Normal-Distribution:CVPR-2013}. However, by this idea, the inherent structure in the problem has not been exploited, which may hinder the success of 3D reconstruction. Even if the method succeeds in obtaining 3D reconstruction, it cannot tell meaningful segmentation of the multiple non-rigid objects.  Another choice would be conducting the tasks of non-rigid motion segmentation \cite{SSC:PAMI-2013} and non-rigid 3D reconstruction \cite{Dai-Li-He:CVPR-2012} successively. However, in this way, the solution of each sub-task does not benefit from the solution of the other sub-task. For example, non-rigid motion segmentation provides critical information to correct non-rigid 3D reconstruction while non-rigid 3D reconstruction constrains the corresponding multi-body motion segmentation. Therefore, we would like to emphasize that since non-rigid deformation originally occurs in 3D space, it's more intuitive to perform non-rigid motion segmentation and reconstruction simultaneously in 3D space rather than solving this problem using two step process.


{\small \begin{table*}[h!]
\caption{A classification of different SFM problems defined by the number of objects and the rigidity of each object. This paper aims to fill in the currently missing work of {\bf Multi-body Non-rigid SFM} shown in blue.\label{tab:summary}}\vspace{-0.15in} 
\center
\resizebox{1.0\linewidth}{!} {\small
\begin{tabular}{|l||c|c|}
\hline
 & Single body & Multi-body\\\hline\hline Rigid &  \begin{tabular}{@{}c@{}} Single-body Rigid SFM \cite{Tomasi:IJCV-1992} \\ $\m W_{2\m F\times \m P} = \m R_{2\m F\times 3} \m S_{3\times \m P}, \rank(\m S) = 3$ \end{tabular}   & \begin{tabular}{@{}c@{}} Multi-body Rigid SFM \cite{Multi-body-Rigid:IJCV-1998}\cite{Yan-Pollefeys:PAMI-2008} \\ $\m W_{2\m F\times \m P} = \m R_{2\m F\times 3\m F} \m S_{3\m F\times \m P}, \rank(\m S) = 3\m K $ \end{tabular}     \\ \hline
Non-rigid & \begin{tabular}{@{}c@{}} Single-body Non-rigid SFM \cite{Bregler:CVPR-2000} \\ $\m W_{2\m F\times \m P} = \m R_{2\m F\times 3\m F} \m S_{3\m F\times \m P}, \rank(\m S) = 3\m K$ \end{tabular}   &  \begin{tabular}{@{}c@{}}  \textcolor{blue}{Multi-body Non-Rigid SFM}\\ \textcolor{blue}{$\m W_{2\m F\times \m P} = \m R_{2\m F\times 3\m F} \m S_{3\m F\times \m P}, \m S = \m S \m C$. i.e, } \\ \textcolor{blue}{3D trajectory should lie in union of linear/affine subspace.} \end{tabular}    \\ \hline
\end{tabular}
}
\label{tab:multi-scal}
\end{table*}}

\begin{figure}
\centering
\subfigure[\label{fig:multi_object3}] {\includegraphics[width=0.40\textwidth]{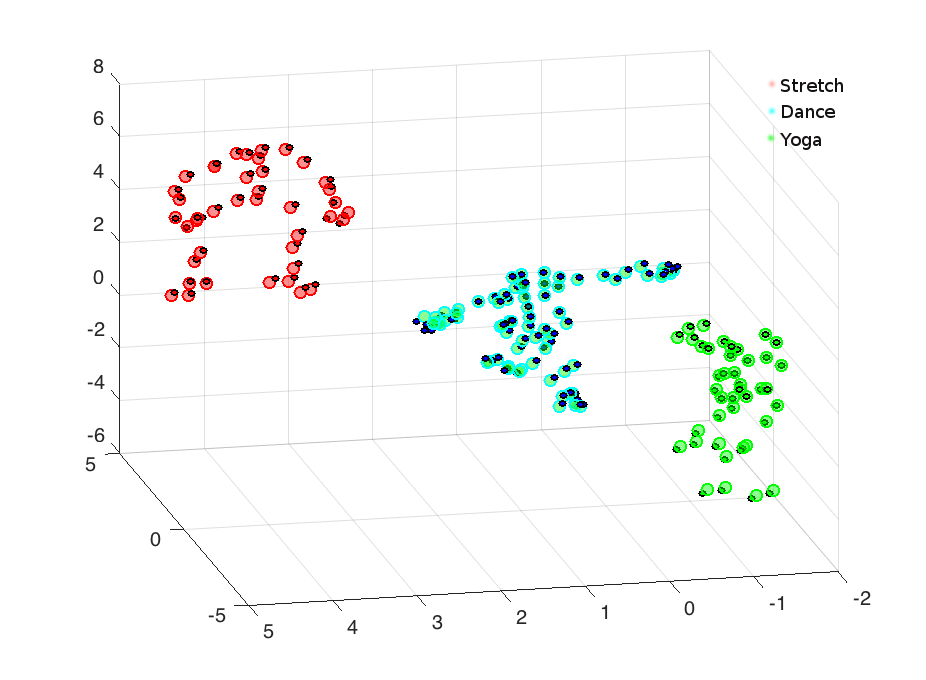}}
\subfigure[\label{fig:diag_multi3}] {\includegraphics[width=0.40\textwidth]{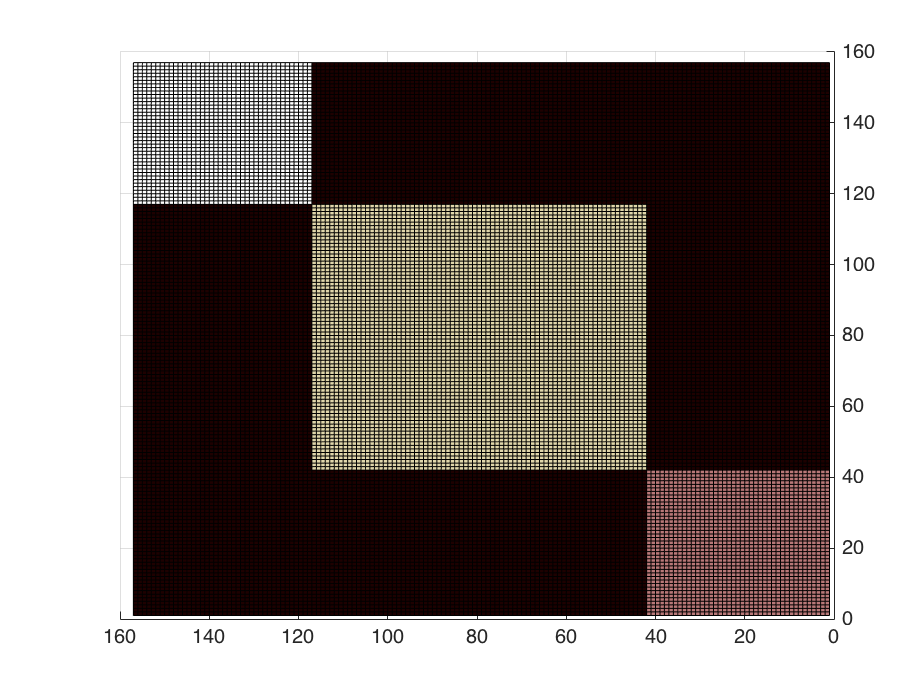}}
\caption{(a) Three subjects are performing actions such as stretch(red), dance(cyan) and yoga(green). Our approach is able to reconstruct and segment each action faithfully with reconstruction error of 0.0413 and 0 segmentation error. Here, different color corresponds to segmentation with dark and light color circles for each subject shows ground-truth and reconstructed 3D coordinates respectively. (b) Obtained block diagonal matrix.}
\noindent\makebox[\linewidth]{\rule{0.83\paperwidth}{0.4pt}}
\label{fig:result_3object}
\end{figure}

To demonstrate the advantage of the concurrent procedure as shown in Fig. \ref{fig:flow_diagram} and Table \ref{tab:summary}, this paper introduces an approach to perform non-rigid 3D reconstruction and its motion segmentation simultaneously. Specifically, we represent the multiple non-rigid motion as union of 3D trajectory subspaces \footnote{Zhu \etal \cite{Union_Subspaces:CVPR-2014} used the union of subspaces representation (different non-rigid deformations lie in different subspaces), where the subspace is defined in shape space contrast to our trajectory space. As we will show later, this difference provides uniqueness of our formulation in dealing with multiple non-rigid deformation and in dealing with dense case.}. By using the self-expressiveness model in representing multiple linear/affine subspace, where each 3D trajectory can be expressed with other trajectories in the same subspace only, enables us in compact representation of trajectories. In this way, we are able to exploit the inherent grouping structure in \emph{3D trajectory space}. For dense non-rigid reconstruction, we could further enforce the spatial coherence constraint. By contrast to existing methods, this endows us the following benefits:
\begin{enumerate}[(i)]
\item A compact representation for component non-rigid deformation in 3D trajectory space;
\item Joint multiple deformable objects motion segmentation and 3D non-rigid structure reconstruction;
\item Improved spatial regularity in 3D non-rigid dense reconstruction (in contrast, a hard segmentation of the 2D tracks may result in discontinuity at the segmentation boundary). \end{enumerate}

The readers are invited to have a preview of results obtained by our method, as shown in Fig.~\ref{fig:multiple_nonrigid}, where we show a daily life motion capture of two person performing dancing and yoga individually as in Fig.~\ref{fig:dance_yoga_2D_1} and Fig.~\ref{fig:dance_yoga_2D_2}. Similarly, in Fig.~\ref{fig:result_3object} three different person are performing stretch, dance and yoga. By applying our proposed method, we are able to achieve both non-rigid 3D reconstruction and non-rigid motion segmentation as shown in Fig.~\ref{fig:dance_yoga_3D_1}, Fig.~\ref{fig:dance_yoga_3D_2} and Fig.~\ref{fig:multi_object3}. 

\paragraph{Contributions:} (1) This paper is first to model multiple non-rigid deformations as points in the union of multiple affine 3D trajectory subspace. This enables us to jointly solve the non-rigid reconstruction and non-rigid motion segmentation problems in 3D trajectory space; (2) Our formulation can handle both sparse and dense multiple non-rigid reconstruction problems uniformly; (3) We propose an efficient optimization procedure based on ADMM.

\section{Related Work}
Ever since the seminal work by Bregler \etal \cite{Bregler:CVPR-2000} modeling a non-rigid shape as lying in a ``\emph{shape space}'' (a linear combination of basis shapes), considerable progress has been made in the area of non-rigid 3D reconstruction. In 2004, Xiao \etal \cite{Xiao-Chai-Kanade:ECCV-2004} showed the inherent ambiguity in modeling non-rigid shape and proposed a remedy of ``basis constraints'' to derive a closed-form solution. In 2008, Akhter \etal \cite{Akhter-Sheikh-Khan-Kanade:Trajectory-Space-2010} presented a dual approach by modeling 3D trajectories, \ie ``\emph{trajectory space}''. In 2009, Akhter \etal \cite{Defence-orthonormality:CVPR-2009} proved that even there is an ambiguity in shape bases or trajectory bases, non-rigid shapes can still be solved uniquely without any ambiguity. Based on this, in 2012, Dai \etal \cite{Dai-Li-He:CVPR-2012} proposed a ``prior-free'' method to recover camera motion and 3D non-rigid deformation by exploiting the low rank constraint only. Besides shape basis model and trajectory basis model, the shape-trajectory approach \cite{Complementary-rank-3:CVPR-2011} combines two models and formulates the problems as revealing trajectory of the shape basis coefficients. Besides linear combination model, Lee \etal \cite{Procrustean-Normal-Distribution:CVPR-2013} proposed a Procrustean Normal Distribution (PND) model, where 3D shapes are aligned and fit into a normal distribution. Simon \etal \cite{Spatiotemporal-Priors:ECCV-2014} proposed to exploit the Kronecker pattern in the shape-trajectory (spatial-temporal) priors. Zhu and Lucey \cite{Convolutional-Sparse-Coding-Trajectory:PAMI-2015} applied the convolutional sparse coding technique to NRSFM using point trajectories. However, the method requires to learn an over-complete basis of 3D trajectories, prior to performing 3D reconstruction.

Despite of the above success, NRSFM is still far behind its rigid counterpart. This is mainly due to  difficulty in modeling real world non-rigid deformation. Real world non-rigid reconstruction generally requires the ability to handle long-term, complex and dense non-rigid shape variations. Such complex and dense non-rigid motion not only increases the computational complexity but also adds difficulty in modeling various kinds of different motions. Zhu \etal \cite{Union_Subspaces:CVPR-2014} proposed to represent long-term complex non-rigid motion as lying in a union of shape sub-spaces rather than sum of sub-spaces. Cho \etal \cite{PND-Mixture-Model:IJCV-2016} represented complex shape variations probabilistically by a mixture of primitive shape variations.

By contrast to the above methods dealing with sparse NRSFM, dense NRSFM methods such as \cite{Dense-NRSFM:3DIMPVT-2012}\cite{Dense-NRSFM:CVPR-2013}\cite{Video-Registration:IJCV-2013}\cite{Video-Pop-Up:ECCV-2014} aim at achieving 3D reconstruction for each pixel in a video sequence, where spatial constraint has been widely exploited to regularize the problem. Garg \etal \cite{Dense-NRSFM:CVPR-2013} presented a variational formulation to dense non-rigid reconstruction by exploiting the spatial smoothness in 3D shapes, which in principle deals with single non-rigid deformation in contrast to our multiple non-rigid deformations. Fragkiadaki \etal \cite{Grouping-Trajectory-Reconstuction:NIPS-2014} solved the problem in sequel, namely, video segmentation by multi-scale trajectory clustering, 2D trajectory completion, rotation estimation and 3D reconstruction. Recently, Yu \etal \cite{Direct-Dense-Deformable:ICCV-2015} bridges template based method and feature track based method by proposing a dense template based direct approach to deformable shape reconstruction from monocular sequences.

Russell \etal \cite{Video-Pop-Up:ECCV-2014} proposed to simultaneously segment a complex dynamic scene containing mixture of multiple objects into constituent objects and reconstruct a 3D model of the scene by formulating the problem as hierarchical graph-cut based segmentation, where the entire scene is decomposed into background and foreground objects and the complex motion of non-rigid or articulated objects are modeled as a set of overlapping rigid parts. Our proposed method differs from this method in the following aspects: 1) We provide a compact representation to multiple non-rigid deformation problem; 2) We propose an efficient and elegant optimization based on ADMM; 3) Our method could deal with both sparse and dense scenarios elegantly.

\section{Formulation}
We seek to reconstruct the 3D trajectories such that they satisfy the union of affine subspace constraint (\ie the 3D trajectories lie in a union of affine subspaces) and non-rigid shape constraints (low rank and spatial coherent).

Let us consider a monocular camera observing multiple non-rigid objects. In this paper, we use the $orthographic$ $camera$ model and eliminate the translation component in camera motion as suggested in \cite{Bregler:CVPR-2000}. The image measurement $\m w_{ij} = [u_{ij}, v_{ij}]^T$ and 3D point $\m S_{ij}$ on the non-rigid shape are related by the camera motion $\m R_i$ as: $\mathbf{w}_{ij} = \m R_i \m{S}_{ij}$, where $\m R_i \in \mathbb{R}^{2\times 3}$ denotes the first two rows of the $i$-th camera rotation. Under this representation, stacking all the $\m F$ frames of measurements and all the $\m P$ points in a matrix form will give us: 
\begin{equation}
\label{eq:motion_shape}
\m W = \m R \m S,
\end{equation}
where $\m R = \mathrm{blkdiag}(\m R_1,\cdots,\m R_F) \in \mathbb{R}^{2\m F\times 3\m F}$ denotes the camera motion matrix. NRSFM aims at recovering the \emph{camera motion} $\m R$ and 3D non-rigid reconstruction $\m S \in \mathbb{R}^{3\m F\times \m P}$ from the 2D \emph{measurement matrix} $\m W \in \mathbb{R}^{2\m F \times \m P}$ such that $\m W=\m R\m S$. 

\subsection{Representing multi-body non-rigid structure as a union of affine subspace}
\begin{figure}[!htp]
  \begin{center}
  \subfigure[\label{fig:affine_s}]{\includegraphics[width=0.32\linewidth]{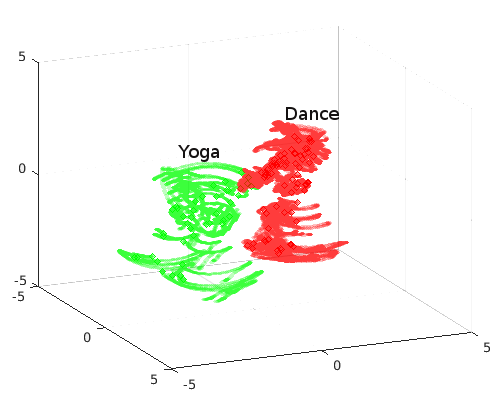}}
  \subfigure[\label{fig:A_algo}]{\includegraphics[width=0.32\linewidth]{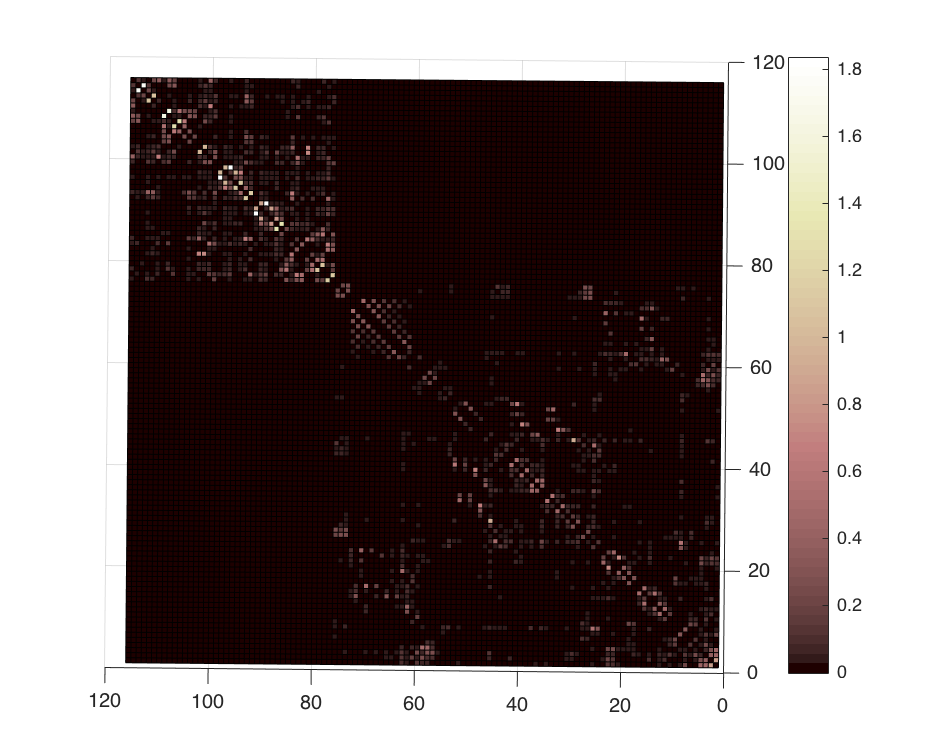}}
  \subfigure[\label{fig:A_clean}]{\includegraphics[width=0.32\linewidth]{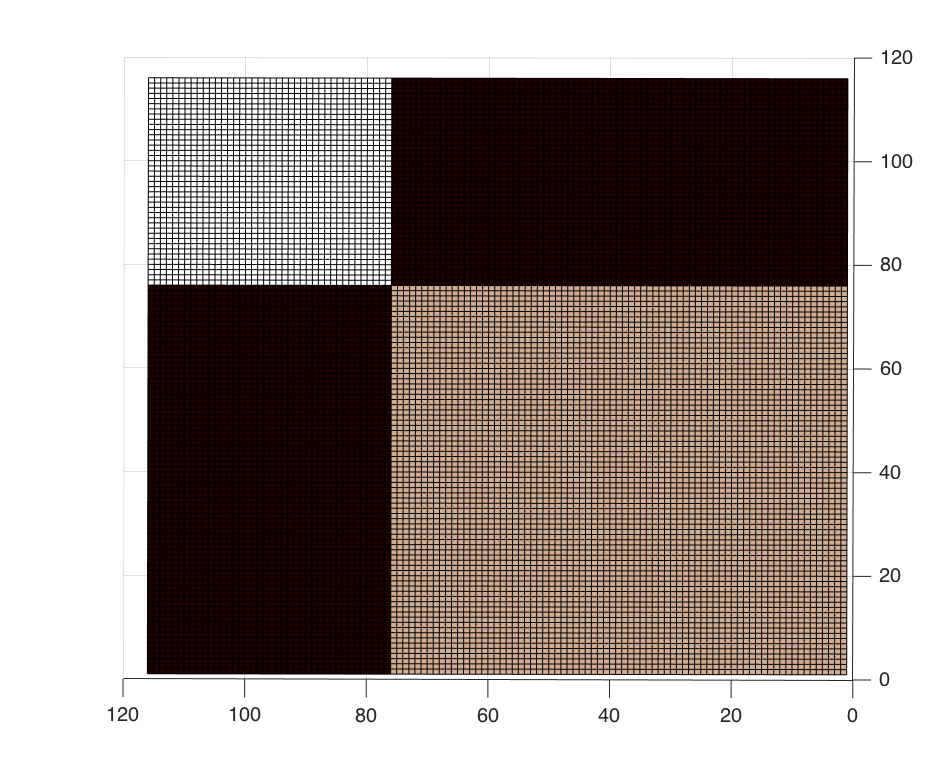}}
  \end{center}
  \caption{\small (a) Two subjects performing different non-rigid motion that are dance and yoga. Red and green color shows the entire trajectory of each objects over $F$ frames. (b) Visualisation of the affinity matrix obtained using our formulation. (c) Clean affinity matrix obtained after incorporating spectral clustering \cite{Spectral-Clustering:NIPS-2001}.  }
  \label{fig:Union of Subspace representation}
  \noindent\makebox[\linewidth]{\rule{0.83\paperwidth}{0.4pt}}
\end{figure}

We propose that multiple non-rigid structures that corresponds to distinct motion lies in a union of affine subspace. Here, the underlying assumption is that the trajectories belong to different non-rigid objects spans a distinct affine subspace. Figure \ref{fig:A_algo} clearly validates such assumption as there are only connection within clusters and no connection between clusters or block diagonal structure. 

Now, consider each trajectory $\m S_i$ that corresponds to a $3 \m F$ dimensional vector formed by stacking the 3D feature tracks of point $i$ across all frames. 
\begin{equation}
\m S_i = [\m S_{1i}^T, \m S_{2i}^T, ..., \m S_{\m Fi}^T]^T \in \mathbb{R}^{3\m F},
\end{equation}
where $\m S_{fi}$ $\in $ $\mathbb{R}^3$. Under the orthographic camera model, feature trajectory associated with non-rigid motion lie in an affine subspace of dimension $\mathbb{R}^{3 \m F}$. After taking $\m P$ such $3\m F$ dimensional trajectory and stacking into the column of a matrix we form $\m S$ matrix $\in $ $\mathbb{R}^{3\m F \times \m P}$. Mathematically, it implies that $\m S$ is a real valued matrix whose columns are drawn from a union of $n$ subspace of $\mathbb{R}^{3\m F}$ of dimension less than min($3\m F$, $\m P$). Therefore, each trajectory in a union of affine subspace could be faithfully reconstructed by a combination of other trajectories in the same subspace. This leads to $self$-$expressiveness$ of 3D trajectories. Concretely, 
\begin{equation}
\m S_i = \m S \m C_i, \m C_{ii} = 0.
\end{equation}

\begin{figure}
\centering
\includegraphics[width=0.6\textwidth] {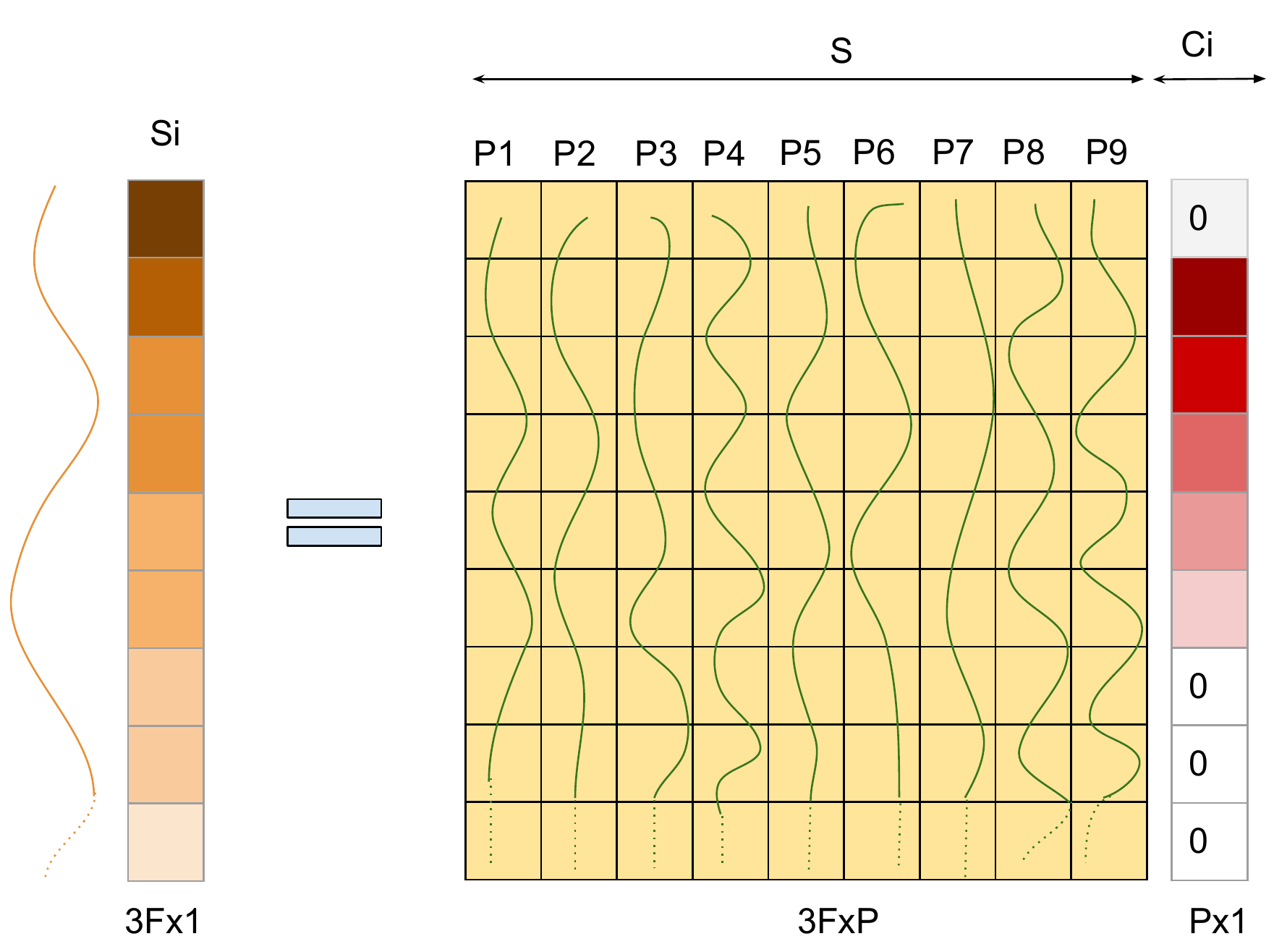}~~~
\caption{Visual illustration of the affine subspace constraint $\m S_i$ = $\m S$ $\m C_i$ for multi-body NRSFM. Each column of $\m S$ is a trajectory of a 3D point(shown in green). This visualization tries to state that a trajectory $\m S_i$ can be reconstructed using affine combination of few other trajectories(sparse). $Note :$ This pictorial representation is provided for better understanding and is only for illustration purpose, signals shown here are not generated mathematically. } 
\label{fig:S_Sci}
\end{figure}

Here, $\m C_i$ is a $\m P \times 1$ coefficient vector and $\m C_{ii}$ = $0$ takes care of the trivial solution $\m S_i = \m S_i$. Stacking all such coefficient vectors, we form a matrix $\m C \in \mathbb{R}^{\m P \times \m P}$ that captures the similarity between different trajectories. Using the fact that any trajectory of $\m S$ in an affine subspace of dimension $3\m F$ can be written as affine combination of $3 \m F + 1$ other points from $\m S$ and to cluster trajectories that lies near to union of affine subspaces, we arrive at the following equation.
\begin{equation}
\label{eq:SSC_constraints}
\m S = \m S \m C, \m 1^T \m C = 1^T,  \mathrm{diag}(\m C) = 0.
\end{equation}
Figure \ref{fig:Union of Subspace representation} shows the affinity matrix $\m A = |\m C| + |\m C^T|$ obtained for two objects that undergo non-rigid deformation. The obtained sparse solution clearly shows that multi-body non-rigid structure can be represented as union of affine subspace.

\subsection{Representing multiple non-rigid deformations in case of sparse feature tracks}
To solve the problem of multi-body NRSFM in case of sparse feature tracks, we propose the following optimization framework for simultaneous reconstruction and segmentation of objects that are undergoing non-rigid deformation,
\begin{equation}
\begin{aligned}
& \displaystyle \underset{\m C, \m S, \m S^{\sharp}} {\text{minimize}} \frac{1}{2}\|\m W - \m R \m S \|_F^2 + \lambda_1\|\m C\|_1 + \lambda_2\|\m S^{\sharp}\|_{*} \\
& \text{{subject to: }} \\
& \displaystyle \m S^{\sharp} = g(\m S), 
\displaystyle \m S = \m S \m C, 
\displaystyle \m 1^{T} \m C = \m 1^{T}, 
\displaystyle \mathrm{diag}(\m C) = \m 0.
\end{aligned}
\label{eq:sparse_version}
\end{equation}

The first term in the above optimization is meant for penalizing reprojection error under $orthographic$ projection. Under single-body NRSFM configuration, 3D shape $\m S$ can be well characterized as lying in a single low dimensional linear subspace. However, when there are multiple non-rigid objects, each non-rigid object could be characterized as lying in an affine subspace. One can argue that affine subspace of dimension $n$ can be considered as a subset of $(n+1)$ dimension that includes origin. Nonetheless, such representation may result in ambiguous solution while clustering different subspace.
The fact that the 3D trajectories lie in a union of affine subspaces as argued previously we put the Eq.~\ref{eq:SSC_constraints} as our optimization constraints. 

In addition to this, to reveal the intrinsic structure of multi-body non-rigid structure-from-motion (NRSFM), we seek for sparsest solution of $\m C$ (\cite{SSC:PAMI-2013}). So, the second term in {\bf{Eq.}} \eqref{eq:sparse_version} enforces $l_1$ norm minimization of $\m C$ matrix. Lastly, we enforce a global shape constraint for compact representation of multi-body non-rigid objects by penalizing the rank of entire non-rigid shape. Similar to \cite{Dai-Li-He:CVPR-2012} and \cite{Dense-NRSFM:CVPR-2013}, we penalize the nuclear norm of reshuffled shape matrix $\m S^{\sharp} \in \mathbb{R}^{\m F\times 3 \m P}$, this is because nuclear norm is known as convex envelope of the rank function. Here, $\m g(\m S)$ denotes  mapping from $\m S$ $\in$ $\mathbb{R}^{3\m F\times \m P}$ to $\m S^\sharp$ $\in$ $\mathbb{R}^{\m F\times 3 \m P}$.


\subsection{Representing multi-body non-rigid deformations in case of dense feature tracks}
When per pixel feature tracks are available, we can enforce spatial regularization ($Markovian$ $assumption$) i.e there will be a high correlation between neighbouring features. To exploit this property, the spatial smoothness can be used as a regularization term to further constrain the non-rigid reconstruction. Garg \etal \cite{Dense-NRSFM:CVPR-2013} proposed to use the total variation of the 3D shape $\|\m S\|_{TV}$. By contrast, we propose to enforce the spatial smoothness constraint on the coefficient matrix $\m C$ directly by using the $l_1$ norm,
\begin{equation}
\label{eq:C_TV}
\sum_{(i,j) \in \mathcal{N}} \|\m C_i - \m C_j\|_1,
\end{equation}
\ie, the total variation of $\m C$. This definition gives us the benefit in solving the problem as proved later. In essence, the total variations of $\m C$ and $\m S$ are correlated. However, it is desirable that $\m C$ matrix must cater the self-expressiveness of the non-rigid shape deformation as compact as possible. So, we incorporate spatial smoothness constraint on coefficient matrix than on shape matrix $\m S$. 

\begin{figure}
\centering
\includegraphics[width=0.6\textwidth] {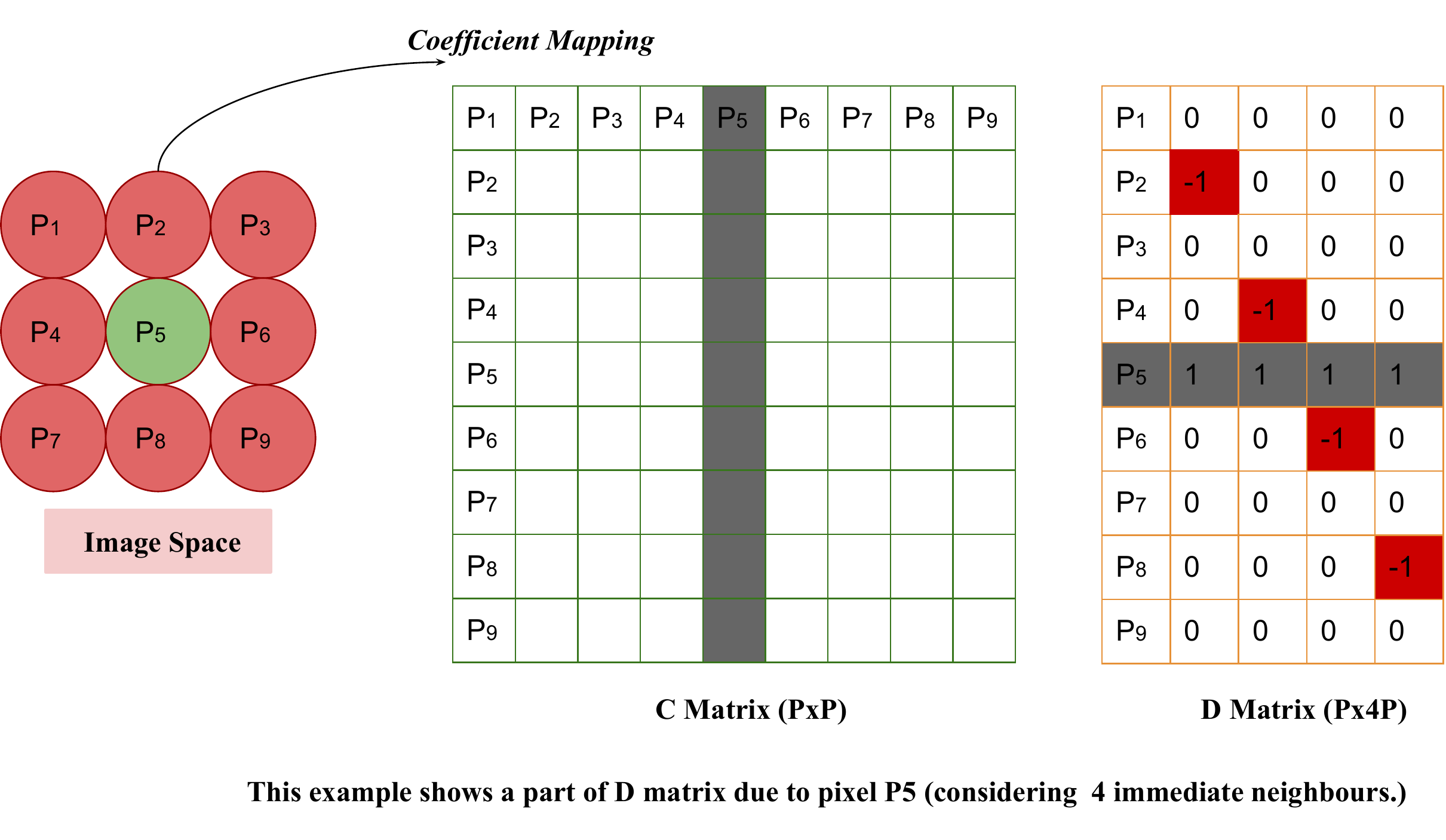}~~~
\caption{D matrix caters the neighboring trajectory relation. In the above illustration $\m P_2$, $\m P_4$, $\m P_6$, $\m P_8$, are the 4 immediate neighboring trajectories of $\m P_5$. Therefore, corresponding elements of $\m D$ matrix has $\m -1$  entries, $\m P_5$ has value $\m 1$ and the rest entries are $\m 0$.  Therefore, the corresponding coefficient column of $\m C$ matrix now rely on relations  defined in the $\m D$ matrix. Here, $\m C$ and $\m D$  are $\m P \times \m P$ and $\m P \times \m 4\m P$ matrix respectively. Where, $\m P$ is the number of total feature tracks. For handling the border pixels one can think of zero padding techniques in 2D image and add zero column vector to the corresponding pixel index in D matrix.} 
\label{fig:DMatrix_exp}
\end{figure}

By introducing an appropriately defined matrix $\m D$ encoding the neighbouring relation, Eq.~\eqref{eq:C_TV} can be equivalently expressed as:
\begin{equation}
\|\m C \m D \|_1 = \sum_{(i,j) \in \mathcal{N}} \|\m C_i - \m C_j\|_1.
\label{eq:C_D}
\end{equation}
In {{Fig.}} \ref{fig:DMatrix_exp}, we illustrate the process of how to obtain the matrix $\m D$.

Adding the spatial constraint to the optimization equation \eqref{eq:sparse_version}, that facilitates this neighboring constraint, we procure the following optimization for dense tracks
\begin{equation}
\begin{aligned}
& \displaystyle \underset{\m C, \m S, \m S^{\sharp} }{\text{minimize}}~ \frac{1}{2}\|\m W - \m R \m S \|_F^2 + \lambda_1\|\m C\|_1 + \lambda_2 \| \m C\m D \|_1 + \lambda_3\|\m S^{\sharp}\|_{*} \\
& \text{{subject to:}} \\
& \displaystyle \m S^{\sharp} = g(\m S), 
\displaystyle \m S = \m S \m C, 
\displaystyle \m 1^{T} \m C = \m 1^{T}, 
\displaystyle \mathrm{diag}(\m C) = \m 0.
\end{aligned}
\label{eq:Optimization_function}
\end{equation}
\section{Solution}
Due to the bilinear term $\m S = \m S \m C$, the overall optimization of Eq.-\eqref{eq:Optimization_function} is non-convex. We solve it via the ADMM, which has a proven effectiveness for many non-convex problems and is widely used in computer vision. The ADMM works by decomposing the original optimization problem into several sub-problems, where each sub-problem can be solved efficiently. To this end, we seek to decompose Eq.-\eqref{eq:Optimization_function} into several sub-problems.

First note that the two $l_1$ terms $\|\m C\|_1$ and $\|\m C \m D\|_1$ can be put together as $\|\m C [\m I ~\m D]\|_1$. Without loss of generality, we still denote the new term as $\m C \m D$, the only difference is, the new dimension of $\m D$ will be $\m P \times 5\m P$, thus the cost function becomes: $\frac{1}{2} \|\m W - \m R \m S \|_F^2 + \lambda_1\|\m C \m D \|_1 + \lambda_2\|\m S^{\sharp}\|_{*} $. To further decouple the constraint, we introduce an auxiliary variable $\m E = \m C \m D$. With these operations, the optimization problem Eq.-\eqref{eq:Optimization_function} can be reformulated as:
\begin{equation}
\begin{aligned}
 & \displaystyle \underset{\m E, \m S, \m S^{\sharp}, \m C} {\text{minimize}} ~\frac{1}{2}\|\m W - \m R \m S \|_F^2 + \lambda_1\|\m E \|_1 + \lambda_2\|\m S^{\sharp}\|_{*} \\
 & \text{{subject to:}} \\
& \displaystyle \m S^{\sharp} =  g(\m S),
\displaystyle \m S = \m S \m C, 
\displaystyle \m C \m D = \m E, 
\displaystyle \m 1^{T} \m C = \m 1^{T}, 
\displaystyle \mathrm{diag}(\m C) = \m 0.
\end{aligned}
\label{eq:main_optimization}
\end{equation}

The Augmented Lagrangian formulation for Eq.-\eqref{eq:main_optimization} is:
\begin{eqnarray}
\mathcal{L}(\m S, \m S^{\sharp}, \m C, \m E, \{\m Y_i\}_{i=1}^4) =  & \frac{1}{2}\|\m W - \m R \m S \|_F^2 + \lambda_1\|\m E \|_1 + \lambda_2\|\m S^{\sharp}\|_{*} + <\m Y_1, \m S^{\sharp} - g(\m S)> \nonumber \\  
&  + \frac{\beta}{2}\|\m S^{\sharp} - g(\m S) \|_F^2 + <\m Y_2, \m S - \m S \m C> + \frac{\beta}{2}\|\m S - \m S \m C \|_F^2  \\ \nonumber & + <\m Y_3,  \m C \m D - \m E> + \frac{\beta}{2}\|\m C \m D - \m E \|_F^2 \nonumber \\ \nonumber & + <\m Y_4, 1^T \m C - \m 1^T > +  \frac{\beta}{2}\|\m 1^T \m C - \m 1^T  \|_F^2.
\label{eq:Lagrangian}
\end{eqnarray}
where $\{\m Y_i\}_{i=1}^4$ are the matrices of Lagrange multipliers corresponding to the four equality constraints, and $\beta$ is the penalty parameter. We do not need to introduce a Lagrange multiplier for the diagonal constraint of $\mathrm{diag}(\m C) = \m 0$ as we will enforce this constraint exactly in the solution of $\m C$.

The ADMM iteratively updates the individual variable so as to minimize $\mathcal{L}$ while the other variables are fixed. In the following, we derive the update for each of the variables.
\subsubsection{The solution of $\m S$:}
\begin{equation}
\begin{array}{l}
\m {S}  =  \arg\min_{\m S} \frac{1}{2}\|\m W - \m R \m S \|_F^2 + <\m Y_1, \m S^{\sharp} -  g(\m S)>  + \frac{\beta}{2}\|\m S^{\sharp} - g(\m S) \|_F^2  \\ \qquad \qquad \qquad + <\m Y_2, \m S - \m S \m C> + \frac{\beta}{2}\|\m S - \m S \m C \|_F^2.  
\end{array}
\end{equation}


The sub-problem for $\m S$ reaches a least squares problem. The closed-form solution of $\m S$ can be derived as:
\begin{equation}
\label{eq:Update_S}
\frac{1}{\beta}(\m R^T \m R + \beta \m I) \m S + \m S(\m I - \m C)(\m I - \m C^T) = \frac{1}{\beta}\m R^T \m W + (g^{-1}(\m S^{\sharp}) + \frac{g^{-1}(\m Y_1)}{\beta } - \frac{\m Y_2}{\beta}(\m I - \m C^T)),
\end{equation}
which is a Sylvester equation.


\subsubsection{The solution of $\m S^{\sharp}$:}
\begin{equation}
\label{eq:update_S_Sharp}
\m {S}^{\sharp} = \arg\min_{\m {S}^{\sharp}} ~ \lambda_2\|\m S^{\sharp}\|_{*} + <\m Y_1, \m S^{\sharp} - g(\m S)>  + \frac{\beta}{2}\|\m S^{\sharp} - g(\m S) \|_F^2
\end{equation}


A close-form solution exists for this sub-problem. Let's define the soft-thresholding operation as $\mathcal{S}_{\tau}[x] = \mathrm{sign}(x)\max(|x|-\tau, 0)$. The optimal solution to Eq.-\eqref{eq:update_S_Sharp} can be obtained as:
\begin{equation}
\label{eq:Update_S_Sharp}
\m S^{\sharp} = \m U \mathcal{S}_{\lambda_2/\beta}(\Sigma) \m V,
\end{equation}
where $[\m U, \Sigma, \m V] = \mathrm{svd}(g(\m S) - \m Y_1/\beta)$.

\subsubsection{The solution of $\m E$:}
\begin{equation}
\m E = \arg\min_{\m E} \lambda_1\|\m E \|_1 + <\m Y_3, \m C \m D -\m E> + \frac{\beta}{2}\| \m C \m D - \m E \|_F^2,  
\end{equation}

A close-form solution exists for this sub-problem by using element-wise shrinkage.
\begin{equation}
\label{eq:Update_E}
\m E = \mathcal{S}_{\lambda_1/\beta}(\m C \m D + \frac{\m Y_3}{\beta}).
\end{equation}

\subsubsection{The solution of $\m C$:}
\begin{equation}
\begin{array}{l}
\m C = \arg\min_{\m C} ~ <\m Y_2, \m S - \m S \m C> + \frac{\beta}{2}\|\m S - \m S \m C\|_F^2 + <\m Y_3, \m C \m D - \m E> + \frac{\beta}{2}\|\m C \m D - \m E\|_F^2 + \\ \qquad \qquad \qquad <\m Y_4, \m 1^T \m C - \m 1^T> + \frac{\beta}{2}\|\m 1^T \m C - \m 1^T \|_F^2 
\end{array}
\end{equation}



The closed-form solution of $\m C$ is derived as:
\begin{equation}\label{eq:update_C}
(\m S^T \m S + \m 1 \m 1^T)\m C + \m C(\m D \m D^T) = \m S^T \m S + \m S^T \frac{\m Y_2}{\beta} + \m E \m D^T - \m Y_3 \frac{\m D^T}{\beta} + \m 1 \m 1^T - \m 1\frac{\m Y_4}{\beta}
\end{equation}
\begin{equation}
\label{eq:Update_C}
\m C = \m C - \mathrm{diag}(\m C),
\end{equation}

Finally, the Lagrange multipliers $\{\m Y_i\}_{i=1}^4$ and $\beta$ are updated as:
\begin{equation}
\label{eq:update_Y1}
\m Y_1 = \m Y_1 + \beta(\m S^{\sharp} - g(\m S)), \m Y_2 = \m Y_2 + \beta(\m S - \m S \m C),
\end{equation}
\begin{equation}
\label{eq:update_Y3}
\m Y_3 = \m Y_3 + \beta(\m C \m D - \m E), \m Y_4 = \m Y_4 + \beta(\m 1^T \m C - \m 1^T),
\end{equation}
\begin{equation}
\label{eq:update_beta}
\beta = \min(\beta_m, \rho\beta),
\end{equation}

\subsubsection{Initialization:} As our method tries to solve a non-convex optimization problem \eqref{eq:main_optimization}, a proper initialization is needed. In this paper, we initialize our implementation by using the single-body non-rigid structure-from-motion method \cite{Dai-Li-He:CVPR-2012}. Other methods can also be used. Note that these single-body non-rigid structure-from-motion was only used for a proper camera motion. In our current implementation, we have fixed the camera motion while updating the 3D non-rigid reconstruction and segmentation. In future, we will put the update of camera rotation in the loop.

\subsubsection{Sparse feature tracks:} In the above section, we provide the derivation for solving the dense feature tracks case. Sparse feature tracks case could be viewed as a simplified form of the dense case by removing the spatial constraint term $\|\m C \m D \|$ directly.


\begin{algorithm}[t!]
\caption{Multi-body non-rigid structure-from-motion and segmentation via the ADMM}
\label{Algorithm 1}
\begin{algorithmic}
\REQUIRE ~~\\
2D feature track matrix $\m W$, camera motion $\m R$, $\lambda_1$, $\lambda_2$, $\rho>1$, $\beta_m$, $\epsilon$; \\ \vspace{0.2cm}
\hspace{-0.3cm}{\bf Initialize:} ${\m S}^{(0)}$, ${\m S^{\sharp}}^{(0)}$, $\m C^{(0)}$, $\m E^{(0)}$, $\{{\bf Y}_i^{(0)}\}_{i=1}^4 = {\bf 0}$, $\beta^{(0)}$;\\ \vspace{0.2cm}
\WHILE {not converged}
\STATE 1. Update $(\m S, \m S^{\sharp}, \m E, \m C)$ by Eq.~\eqref{eq:Update_S}, Eq.~\eqref{eq:Update_S_Sharp}, Eq.~\eqref{eq:Update_E}, Eq.~\eqref{eq:update_C} and Eq.~\eqref{eq:Update_C};\\
\STATE 2. Update $\{{\bf Y}_i\}_{i=1}^4$ and $\beta$ by Eq.~\eqref{eq:update_Y1}-Eq.~\eqref{eq:update_beta};\\
\STATE 3. Check the convergence conditions $\| \m S^{\sharp} - g(\m S)\|_{\infty} \leq \epsilon$, $\| \m S - \m S\m C\|_{\infty} \leq \epsilon$, $\| \m 1^T \m C - \m 1^T \|_{\infty} \leq \epsilon$, and $\| \m C \m D - \m E \|_{\infty} \leq \epsilon$; \\
\ENDWHILE
\vspace{0.2cm}
\ENSURE ~${\m C}$, $\m S, \m S^{\sharp}$.
\STATE Form an affinity matrix $\m A = |\m C| + |\m C^T|$, then apply spectral clustering \cite{Spectral-Clustering:NIPS-2001} to $\m A$. 

Note: Spectral clustering demands number of clusters as input. One way to obtain optimal number of cluster is to use eigen gap information of Laplacian matrix, but in our experiments, we assumed that number of deformable objects are known a priori. 
\end{algorithmic}
\end{algorithm}

\section{Experiments}
To evaluate the effectiveness of our approach for multi-body non-rigid structure-from-motion, we conducted extensive experiments on both synthetic data and real images, under both sparse and dense scenarios. Our implementation is not trying to alternate between shape recovery and SSC \cite{SSC:PAMI-2013}, rather its a unified optimization framework.

As our method jointly perform non-rigid 3D reconstruction and segmentation, we use the following criteria to measure the performances of the algorithm: \\
(i) Relative error in multi-body non-rigid 3D reconstruction 
\begin{equation}
e_{3D} = \|\m S_f^{est} - \m S_f^{GT}\|_F/\| \m S_f^{GT} \|_F,
\end{equation}
(ii) Error in multi-body non-rigid motion segmentation,
\begin{equation}
e_{MS} = \frac{\mathrm{total ~number ~of ~incorrectly ~segmented ~trajectories}}{\mathrm{total ~number ~of trajectories}}.
\end{equation}

\subsection{Multi-body non-rigid data}
\begin{figure}
\centering
\subfigure [\label{fig:1}] {\includegraphics[width=0.24\textwidth]{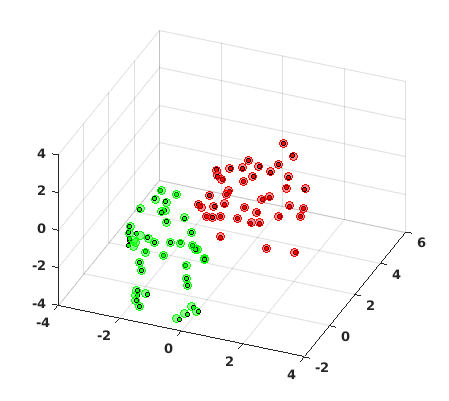}}
\subfigure [\label{fig:2}] {\includegraphics[width=0.24\textwidth]{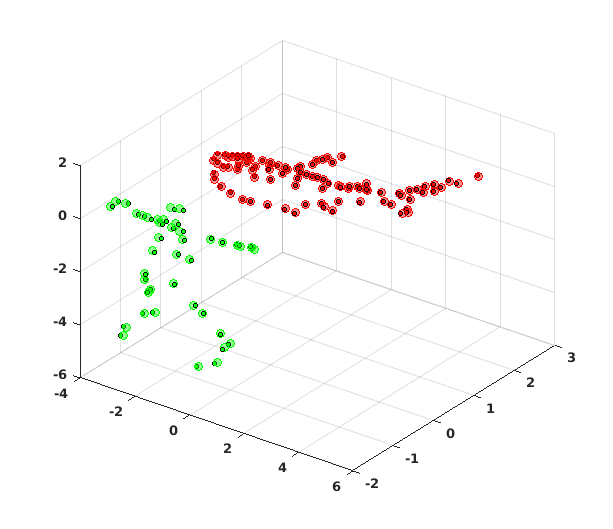}}
\subfigure [\label{fig:3}]{\includegraphics[width=0.24\textwidth]{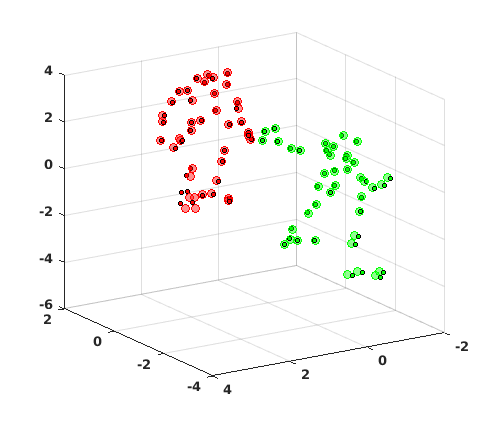}}
\subfigure [\label{fig:4}]{\includegraphics[width=0.24\textwidth]{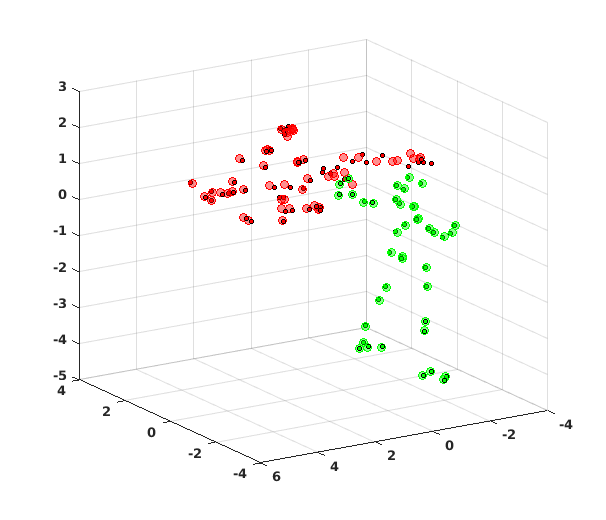}}
\caption{3D reconstruction and segmentation of different multi-body non-rigid motion sequences a) Face-Pickup Sequence; b) Shark-Yoga Sequence; c) Stretch-Yoga Sequence; d) Walking-Yoga Sequence. The dark small circles in the respective segments shows the Ground-Truth 3D points.}
\label{fig:synthetic_sequence}
\noindent\makebox[\linewidth]{\rule{0.83\paperwidth}{0.4pt}}
\end{figure}

To prove the correctness of our formulation, we first test our method on synthetic dataset. We synthesize two non-rigid objects by using the CMU Mocap dataset \cite{Akhter-Sheikh-Khan-Kanade:Trajectory-Space-2010}, for example one person performs Yoga while the other person is dancing. In Fig.~\ref{fig:synthetic_sequence}, we illustrate four sample multi-body non-rigid structure-from-motion sequences. 


\subsection{Experiment 1: Convergence}  
\begin{figure}
\centering
\includegraphics[width=1.0\textwidth,height=0.5\textwidth] {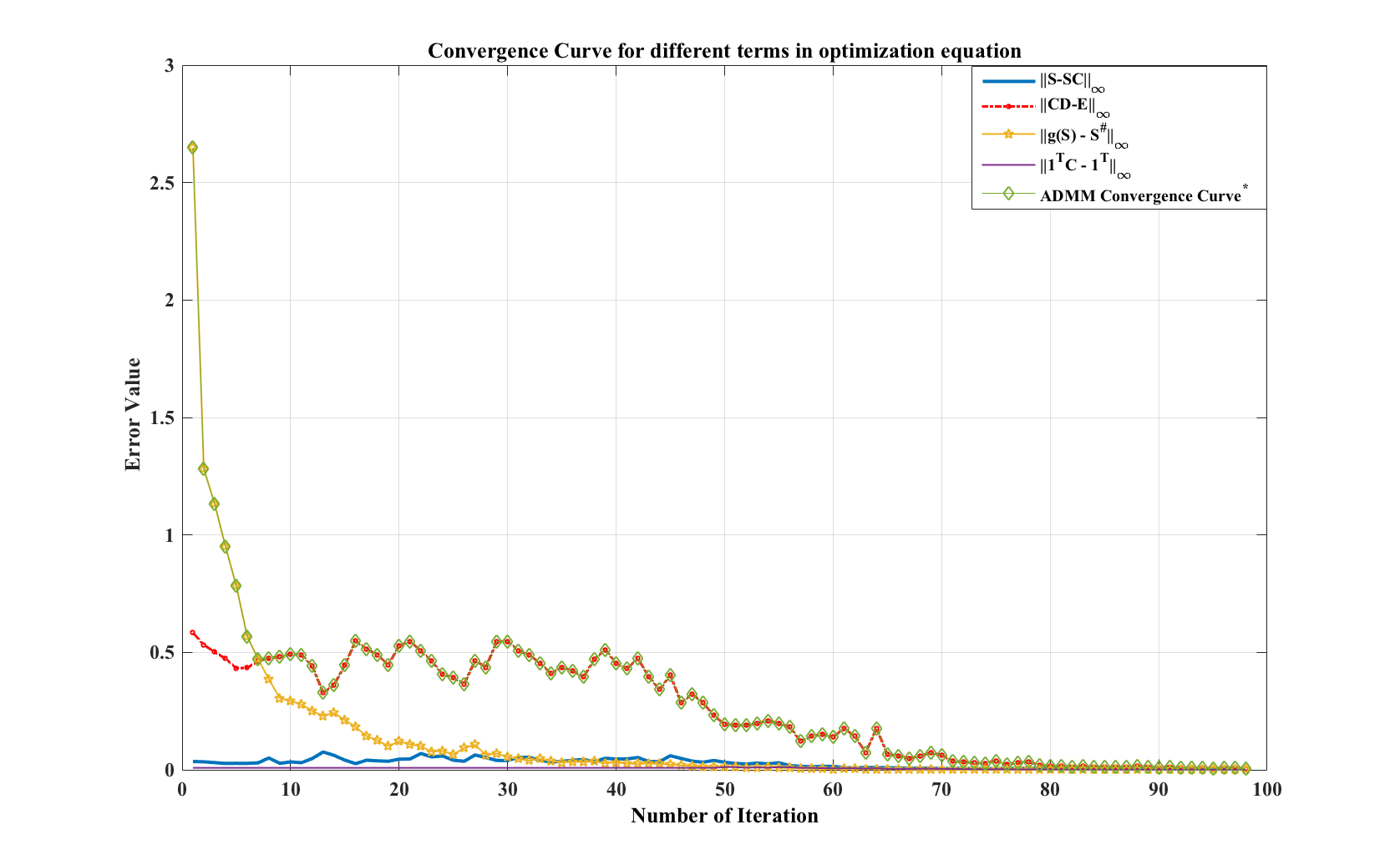}~~~
\caption{Typical convergence curves of the objective function and the primal residuals $\|\m S^{\sharp} - g(\m S)\|_\infty, \| \m S - \m S \m C \|_\infty, \| \m C \m D - \m E\|_\infty$ and $\| \m 1^T \m C - \m 1^T\|_\infty$. The above plot shows the convergence statistics for Dance+Yoga Sequence. *ADMM Convergence Curve = maximum of ($\|\m S^{\sharp} - g(\m S)\|_\infty, \| \m S - \m S \m C \|_\infty, \| \m C \m D - \m E\|_\infty$ and $\| \m 1^T \m C - \m 1^T\|_\infty$). }
\label{fig:Convergence_Curve}
\end{figure}

Given noise free input, we want to check whether or not our proposed algorithm converge; and if it does converge, whether it converges to the correct solution. Note that we use the sparse sequences from the CMU MoCap dataset \cite{Akhter-Sheikh-Khan-Kanade:Trajectory-Space-2010} directly without any dimension reduction or projection. Typical convergence curves of the objective function and the primal residuals are illustrated in Fig.~\ref{fig:Convergence_Curve}. 


\subsection{Experiment 2: Performance on noisy feature tracks}
We also conducted experiments to analyze the performance of our method under different level of noise on input track features. In the same manner as above, we generated multi-body non-rigid sequences (``Dance + Yoga'', ``Face + Pickup'', ``Face + Yoga'', ``Shark + Stretch'', ``Shark + Yoga'', ``Stretch + Yoga'' and ``Walking + Yoga''), then zero-mean Gaussian noise with standard deviation $\sigma$ were added to the feature tracks. For each noisy input, we ran our code for 5 times and recorded the mean 3D reconstruction error and non-rigid motion segmentation error.

\begin{figure}
\centering
\includegraphics[width=0.47\textwidth,height=0.166\textheight] {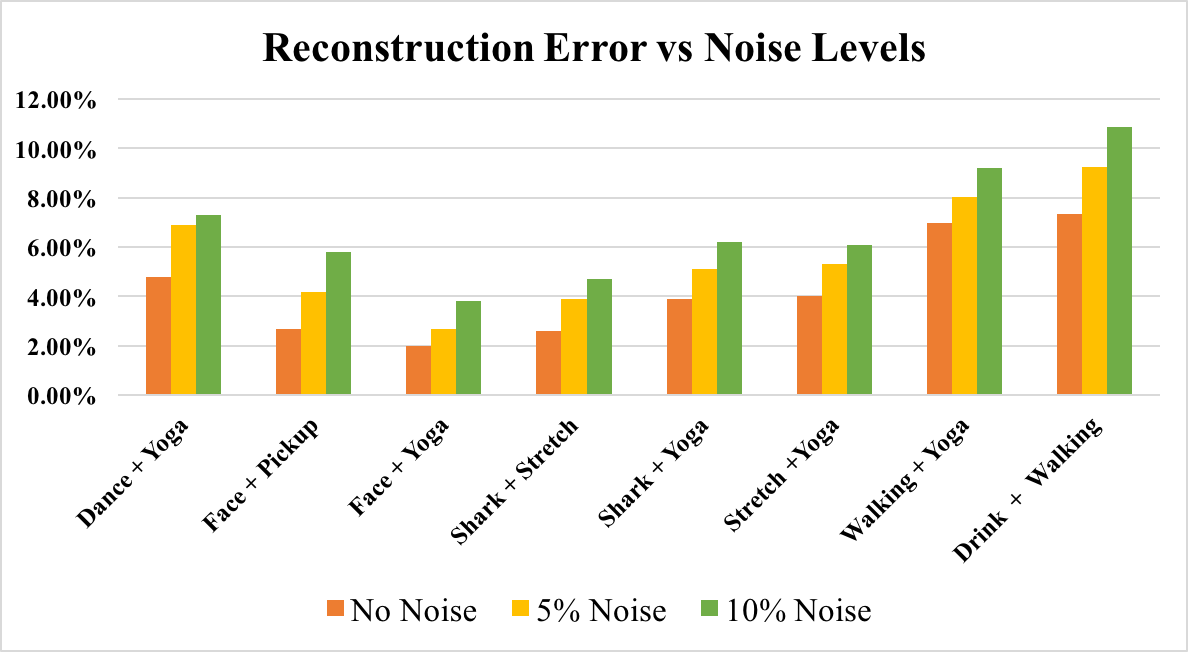}~
\includegraphics[width=0.47\textwidth,height=0.166\textheight]{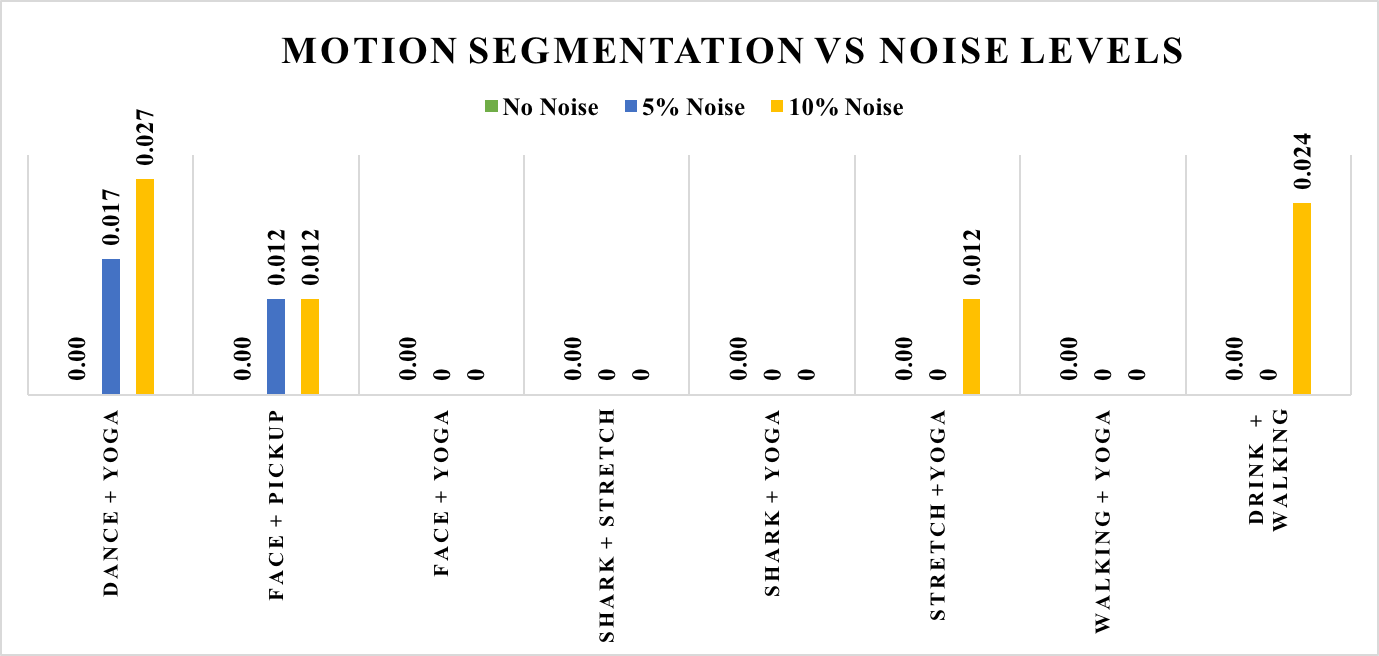}
\caption{Left: 3D Reconstruction error VS noise levels; Right: non-rigid motion segmentation error VS noise levels. }
\label{fig:noise_performance}
\end{figure}

Fig.~\ref{fig:noise_performance}, illustrate the statistical results of 3D non-rigid reconstruction and non-rigid motion segmentation. From the statistics as plotted in Fig.~\ref{fig:noise_performance}, we conclude that both the 3D reconstruction error and the motion segmentation error increases with the increase of noise level. Our 3D reconstruction based non-rigid motion segmentation achieves smaller motion segmentation error compared with 2D trajectory based motion segmentation methods such as sparse subspace clustering (SSC) \cite{SSC:PAMI-2013} and efficient dense subspace clustering (EDSC) \cite{EDSC:WACV-2014}.

\subsection{Experiment 3: Performance Comparison on Sparse Dataset} In Table~\ref{tab:segmentation_comparison} we compared the segmentation results of our approach with SSC \cite{SSC:PAMI-2013} and EDSC \cite{EDSC:WACV-2014} on multi-body non-rigid sequences over 2D feature tracks. It clearly demonstrate that performing non-rigid motion segmentation in 3D space using our approach leads to remarkable results. Since, our method jointly solves 3D reconstruction and multi-body non-rigid motion segmentation, we compare our method with the two stage method, namely

\begin{enumerate}[1)]
\item Baseline method 1: Single body non-rigid structure-from-motion (State-of-the-art ``block-matrix method'' \cite{Dai-Li-He:CVPR-2012} was used) followed by subspace clustering of the 3D trajectories (SSC \cite{SSC:PAMI-2013} was used), denoted as ``BMM+SSC(3D)'';
\item Baseline method 2: Subspace clustering of the 2D feature tracks (2D trajectories) followed by single body non-rigid structure-from-motion for each cluster of 2D feature tracks, denoted as ``SSC(2D)+BMM''.
\end{enumerate}

Table~\ref{tab:comparisons_baseline}, provides experimental comparisons between our method and two baseline methods in dealing with multi-body non-rigid structure-from-motion. In all the sequences, our method achieves zero multi-body non-rigid motion segmentation error and comparable 3D non-rigid reconstruction.

\begin{figure}[!htp]
  \begin{center}
  \subfigure[\label{fig:NRSFM_3D}]{\includegraphics[width=0.32\linewidth]{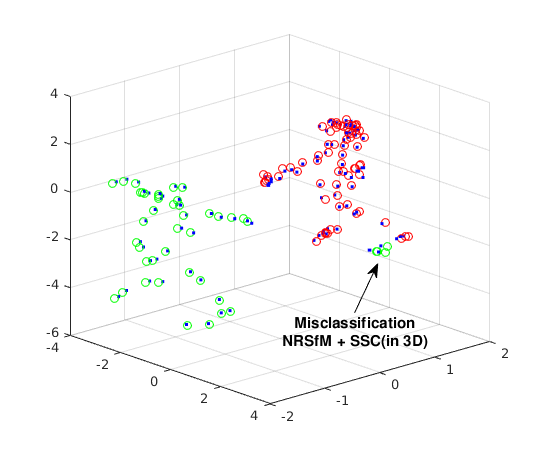}}
  \subfigure[\label{fig:SSC_NRSFM}]{\includegraphics[width=0.32\linewidth]{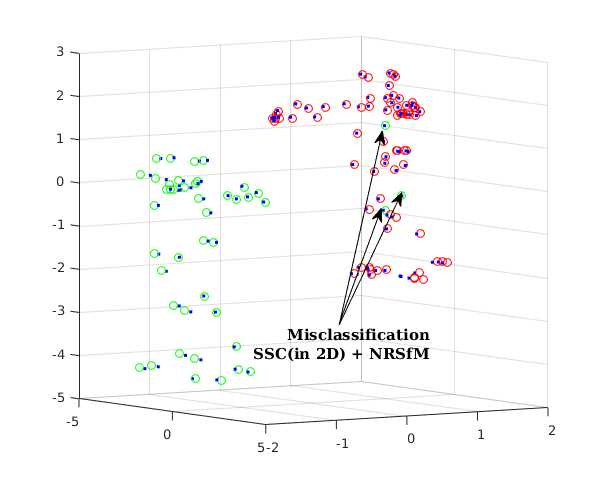}}
  \subfigure[\label{fig:our_method}]{\includegraphics[width=0.32\linewidth]{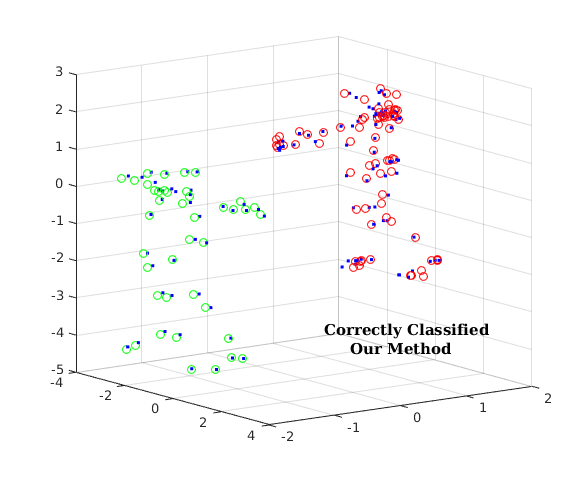}}
  \end{center}
  \caption{\small Demonstrating the efficacy of our approach.  The above plot shows the results on Dance + Yoga sequence. (a) Result obtained by applying BMM method \cite{Dai-Li-He:CVPR-2012} to get 3D points and then use SSC \cite{SSC:PAMI-2013} to segment 3D points. (b)  Result obtained by applying SSC \cite{SSC:PAMI-2013} to 2D feature tracks and then use BMM \cite{Dai-Li-He:CVPR-2012} separately to each segment to get 3D reconstruction. (c) Result by applying simultaneous reconstruction and segmentation framework(Our approach). }
  \label{fig:comparison_efficacy}
  \noindent\makebox[\linewidth]{\rule{0.83\paperwidth}{0.4pt}}
\end{figure}

\begin{figure}[!htp]
  \begin{center}
 \subfigure[\label{fig:shark_yoga}]{\includegraphics[width=0.24\linewidth]{Figures/Shark_Yoga_Affine_Subspace.png}}
 \subfigure[\label{fig:stretch_yoga}]{\includegraphics[width=0.24\linewidth]{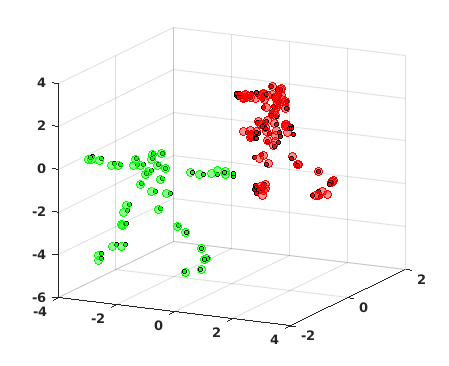}}
 \subfigure[\label{fig:face_yoga}]{\includegraphics[width=0.24\linewidth]{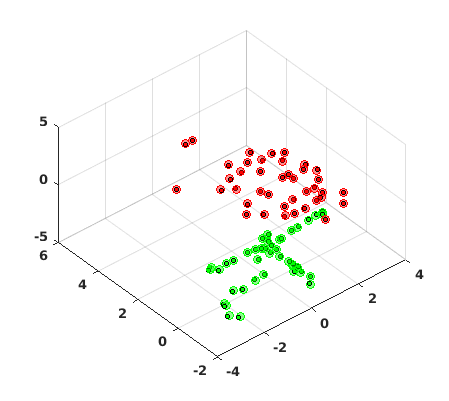}}
 \subfigure[\label{fig:shark_stretch}]{\includegraphics[width=0.24\linewidth]{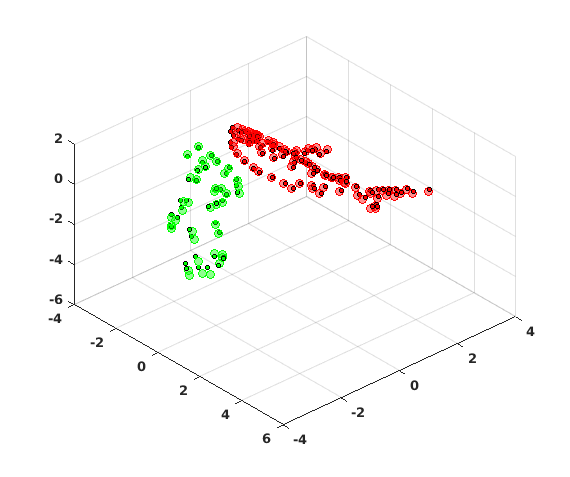}}  
  \end{center}
  \caption{\small Illustration of multi-body non-rigid reconstruction with segmentation on different data-sets. (a) Shark + Yoga Sequence (b) Dance + Yoga Sequence (c) Face + Yoga Sequence (d) Shark + Stretch Sequence. The dark small circles in the respective segments shows the Ground-Truth 3D points.}
  \label{fig:multi_nonrigid}
  \noindent\makebox[\linewidth]{\rule{0.83\paperwidth}{0.4pt}}
\end{figure}

\begin{figure}
\centering
\includegraphics[width=0.6\textwidth, height=0.15\textheight] {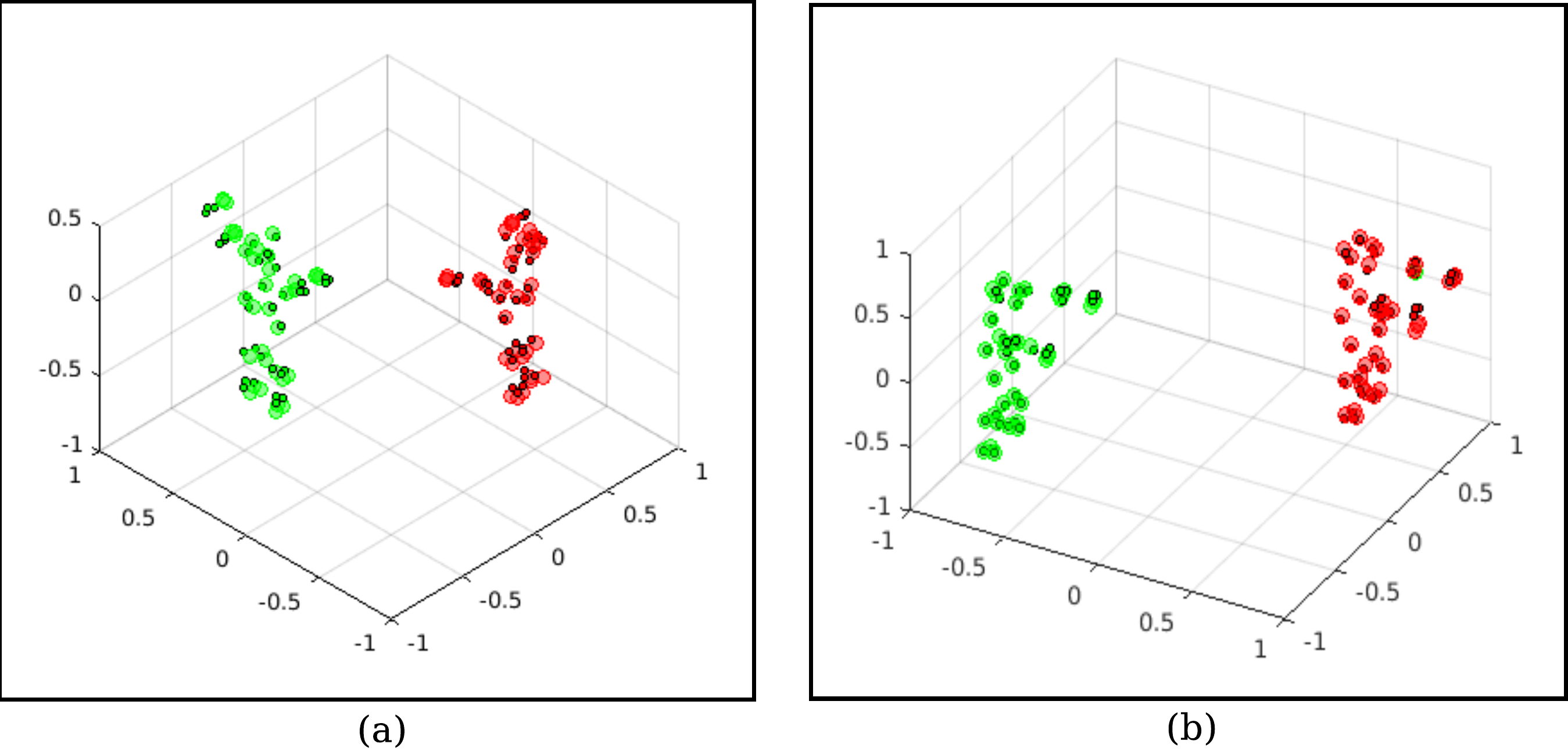}~~~
\caption{Reconstruction and segmentation result on (a) p3\_ball\_1.  (b) p4\_meet\_12 data-set from UMPM database \cite{UMPM}. Light green and light red circles shows the 3D reconstruction, where as dark circles of the respective color shows their 3D ground-truth. Here, different color shows the corresponding segmentation.}
\label{fig:umpm_result}
\noindent\makebox[\linewidth]{\rule{0.83\paperwidth}{0.4pt}}
\end{figure}

Experiments clearly shows that our 3D reconstruction is very close to the accuracy of BMM \cite{Dai-Li-He:CVPR-2012} method in almost all dataset. However, the advantage with our framework we can have robust segmentation along with better reconstruction at the same time. Fig. \ref{fig:comparison_efficacy} shows the robustness of our approach. Our method faithfully reconstruct and segment two different complex non-rigid motion than the two baseline methods. Extensive experiments were performed on synthetic sparse data-sets with different combination of non-rigid motion Fig. \ref{fig:multi_nonrigid} shows some of the results on these different combinations of non-rigid motion on CMU Mocap dataset \cite{Akhter-Sheikh-Khan-Kanade:Trajectory-Space-2010}, where as Fig. \ref{fig:umpm_result} and Table \ref{tab:result_umpm} presents results obtained on UMPM dataset \cite{UMPM}.

\begin{figure}
\centering
\includegraphics[width=0.25\textwidth]{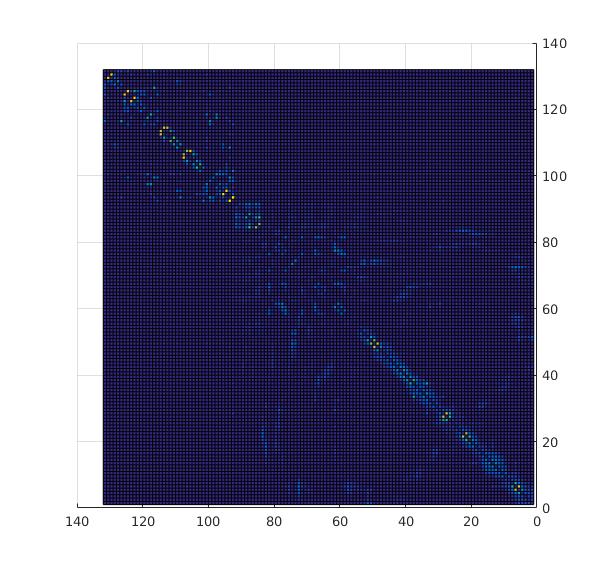}~~~
\includegraphics[width=0.25\textwidth]{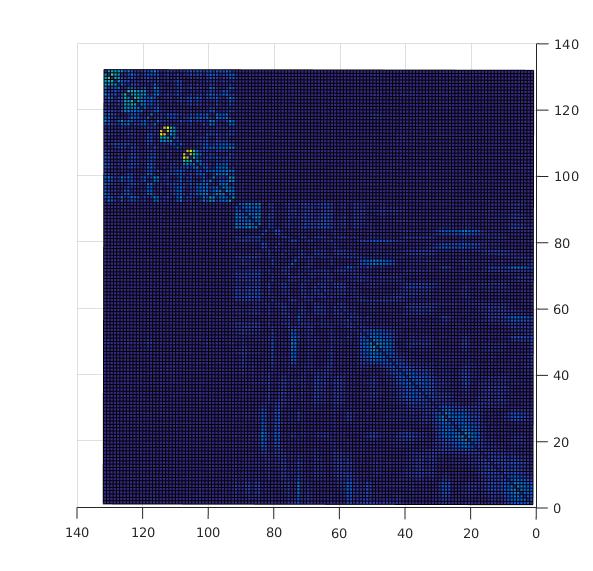}
\caption{Obtained Affinity Matrix $\m A = |\m C| + |\m C^T|$. a) Affinity matrix from SSC; b) Affinity matrix from our Method. Best Viewed on Screen.}
\label{fig:affinity_matrix}
\end{figure}

We also compared the affinity matrices  of our method with SSC \cite{SSC:PAMI-2013}. It's visually apparent from Fig.~\ref{fig:affinity_matrix}, that our method outputs an affinity matrix with better structure, which results in better non-rigid motion segmentation performance.

\begin{table}
\begin{center}
\caption{Motion segmentation performance comparison with SSC \cite{SSC:PAMI-2013} and EDSC \cite{EDSC:WACV-2014} over 2D feature tracks.}
\label{tab:segmentation_comparison}
\begin{tabular}{|c|c|c|c|}
\hline
Dataset & SSC ($e_{MS}$) & EDSC ($e_{MS}$) & Ours\\
\hline
Dance+Yoga & 0.025 & 0.0345 & 0.0\\
\hline
Drink+Walking & 0.0 & 0.01 & 0.0\\
\hline
Shark+Stretch & 0.3939  & 0.0 & 0.0 \\
\hline
Walking+Yoga & 0.0  & 0.0 & 0.0 \\
\hline
Face+Pickup & 0.098 & 0.0 & 0.0\\
\hline
Face+Yoga & 0.012 & 0.0 & 0.0\\
\hline
Shark+Yoga & 0.41 & 0.0 & 0.0 \\
\hline
Stretch+Yoga & 0.0 & 0.0 & 0.0\\
\hline
\end{tabular}
\end{center}
\end{table}

\begin{table}[!htp]
\centering
  \caption{Performance comparison between our method and the baseline methods, where 3D reconstruction error ($e_{3D}$) and non-rigid motion segmentation error ($e_{MS}$) are used as error metrics.}
  \begin{tabular}{|c|c|c|c|c|c|c|}
    \hline
    \multirow{2}{*}{Dataset} & 
      \multicolumn{2}{c|}{BMM + SSC (3D)} &
      \multicolumn{2}{c|}{SSC(2D) + BMM} &
      \multicolumn{2}{c|}{Our Method} \\ \cline{2-7}
    & $e_{3D}$ & $e_{MS}$ & $e_{3D}$ & $e_{MS}$ & $e_{3D}$ & $e_{MS}$ \\
    \hline
    Dance + Yoga & 0.0456 & 0.0345 & 0.0588 & 0.0259 & 0.046 & {\bf{0.0}} \\
    \hline
    Drink + Walking & 0.0745 & 0.0 & 0.0858 & 0.0 & {\bf{0.073}} & {\bf{0.0}} \\
    \hline
    Shark + Stretch & 0.0246 & 0.4015 & 0.0979 & 0.3939 & 0.025 & {\bf{0.0}} \\
    \hline
    Walking + Yoga & 0.0702 & 0.0 & 0.0900 & 0.0 & {0.0702} & {\bf{0.0}} \\
    \hline
    Face + Pickup & 0.0324 & 0.0988 & 0.0239 & 0.0988 & 0.025 & {\bf{0.0}} \\
    \hline
    Face + Yoga & 0.0172 & 0.012 & 0.0332 & 0.012 & 0.019 & {\bf{0.0}} \\
    \hline
    Shark + Yoga & 0.0356 & 0.4167 & 0.1049 & 0.4091 & 0.0371 & {\bf{0.0}}\\
    \hline
    Stretch + Yoga & 0.0392 & 0.0 & 0.0557 & 0.0 & 0.0393 & {\bf{0.0}} \\
    \hline
  \end{tabular}
  \label{tab:comparisons_baseline}
\end{table}

\begin{table}
\centering
  \caption{Quantitative results on UMPM data-set \cite{UMPM}. The rotation were obtained using BMM method \cite{Dai-Li-He:CVPR-2012} (K = 11). Different K value may results in different reconstruction error ($e _{MS}$) value. Statistics clearly reveals that we can get high fidelity motion segmentation using our approach and reasonable 3D reconstruction simultaneously on very complex data-set as well.}
  \begin{tabular}{|c|c|c|c|c|c|c|}
    \hline
    \multirow{2}{*}{Dataset} & 
      \multicolumn{2}{c|}{BMM + SSC (3D)} &
      \multicolumn{2}{c|}{SSC(2D) + BMM} &
      \multicolumn{2}{c|}{Our Method} \\ \cline{2-7}
    & $e_{3D}$ & $e_{MS}$ & $e_{3D}$ & $e_{MS}$ & $e_{3D}$ & $e_{MS}$ \\
    \hline
    p2\_free\_2 & 0.1973 & 0.0 & 0.2177 & 0.0 & 0.1992 & {0.0} \\
    \hline
    p3\_ball\_1 & 0.1356 & 0.0 & 0.1422 & 0.0 & 0.1348 & 0.0 \\
    \hline
    p4\_grab\_2 & 0.2018 & 0.0 & 0.2182 & 0.0 & 0.2080 & 0.0 \\
    \hline
    p4\_table\_12 & 0.2313 & 0.0 & 0.2418 & 0.0 & 0.2313 & 0.0 \\
    \hline
    p4\_meet\_12 & 0.0800 & 0.0135 & 0.0941 & 0.0 & 0.0821 & 0.0135 \\
    \hline
  \end{tabular}
  \label{tab:result_umpm}
\end{table}

\subsection{Experiment 4: Analysis on Dense Dataset}
\begin{figure}[t!]
\centering
\includegraphics[width=0.8\textwidth, height=0.20\textheight] {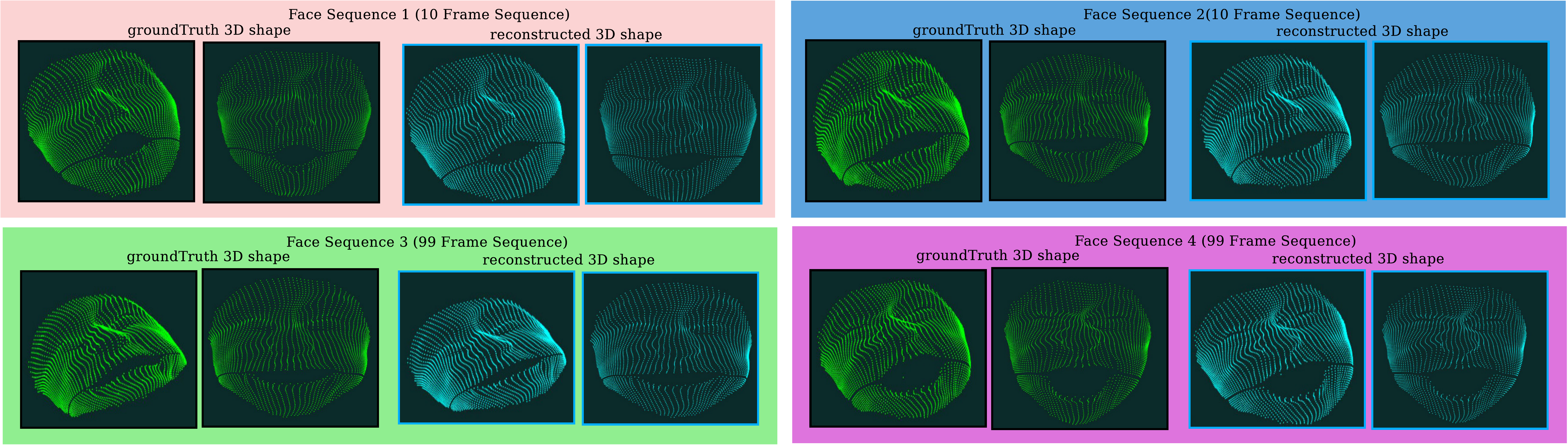}~~~
\caption{Reconstruction results on the synthetic Face Sequence benchmark data-set \cite{Dense-NRSFM:CVPR-2013}. The above shown results were obtained on uniformly sampled trajectories of Face Sequence. }
\label{fig:Dense_result_synthetic}
\end{figure}

\begin{figure}[t!]
\centering
\includegraphics[width=0.8\textwidth, height=0.22\textheight] {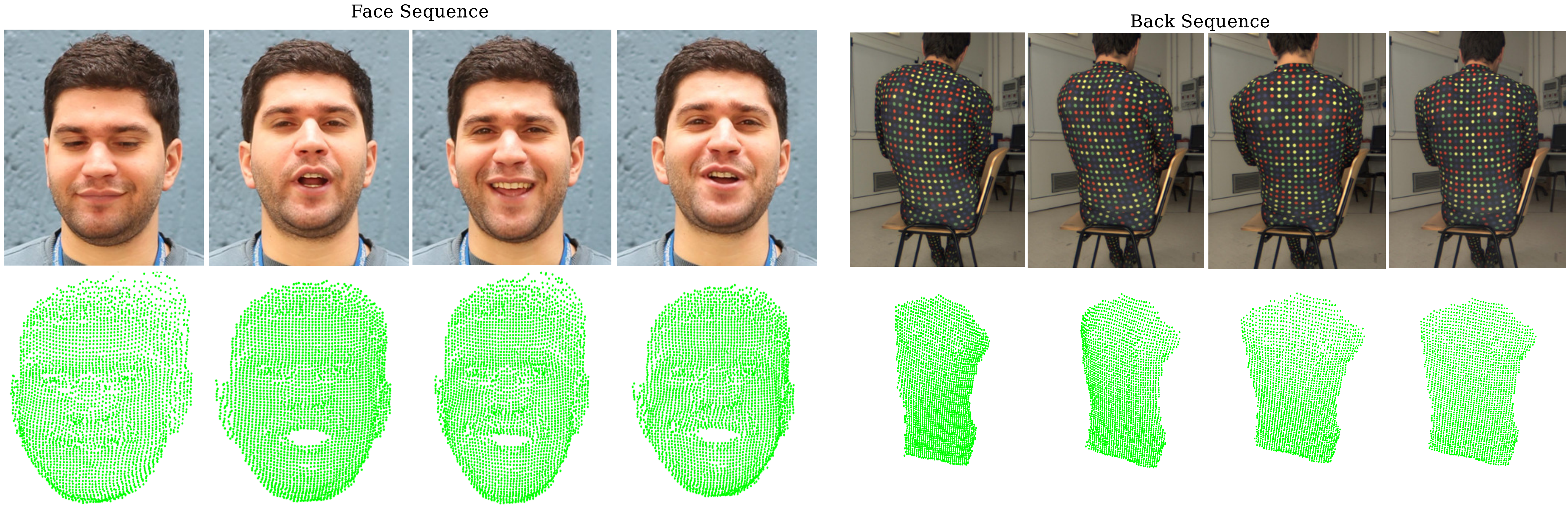}~~~\\
\includegraphics[width=0.8\textwidth, height=0.22\textheight] {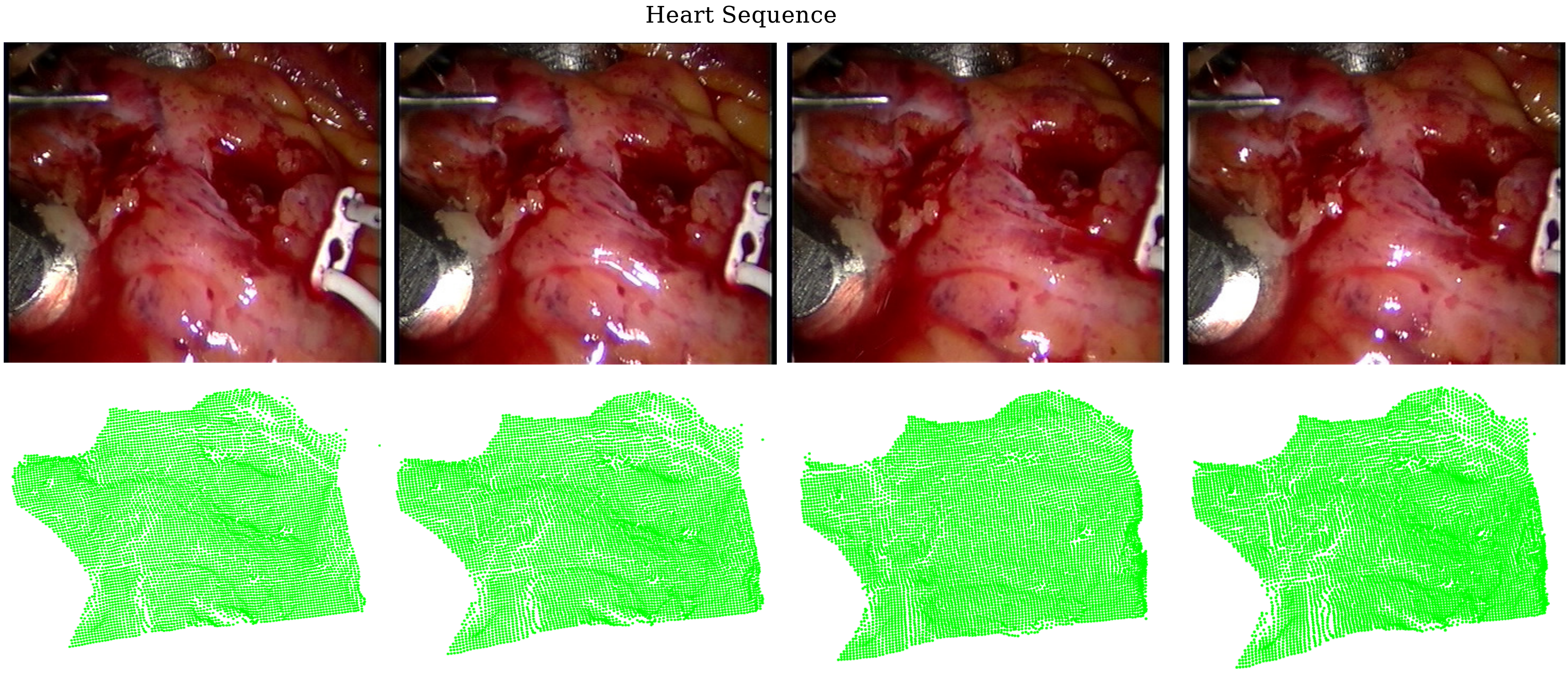}
\caption{Reconstruction results on the real sequence data-sets: Face, Back and Heart Sequence \cite{Dense-NRSFM:CVPR-2013}. The above shown results were obtained on uniformly sampled trajectories.}
\label{fig:real_face_back_}
\end{figure}

\begin{figure}[t!]
\centering
\includegraphics[width=0.8\textwidth, height=0.20\textheight] {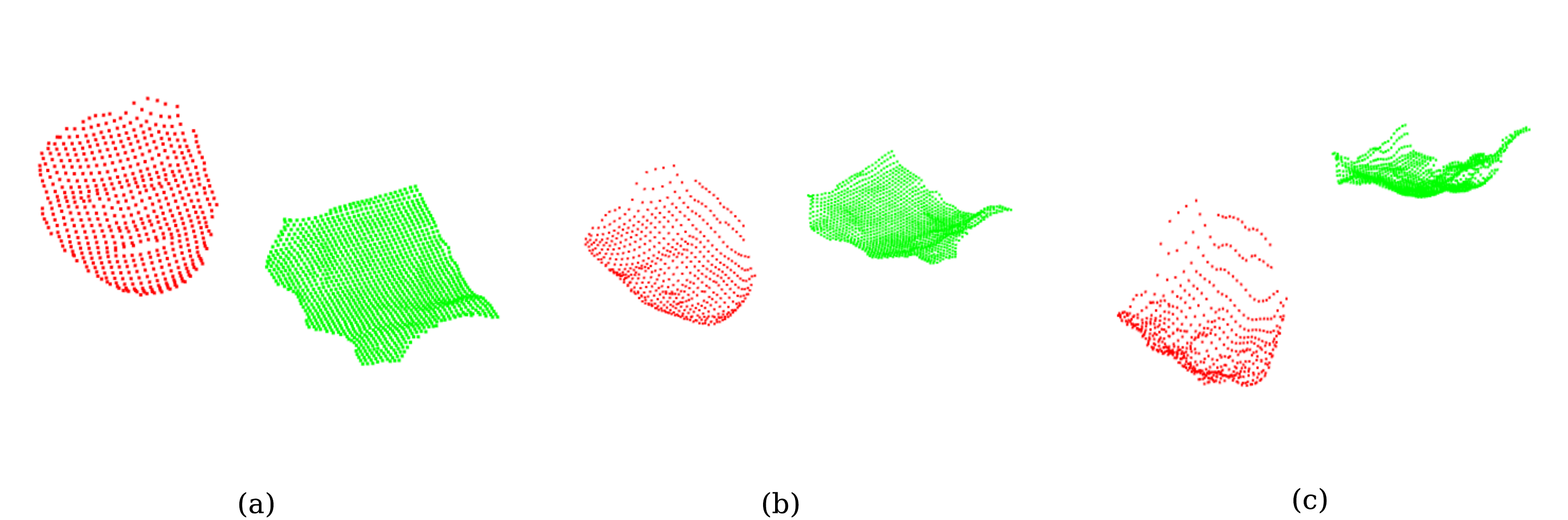}~~~
\caption{Segmentation result on the real sequence data-sets: (Face + Heart) Sequence \cite{Dense-NRSFM:CVPR-2013}. (a) Top-View, (b) \& (c) Side-View of the scene. Different color signature (Green for heart and Red for Face) symbolizes the corresponding class labels.}
\label{fig:real_face_heart_}
\end{figure}

To expeditiously test the effectiveness of our implementation over available dense sequences, we tested our method on the uniformly sampled version of the original sequences. We performed experimentation on benchmark NRSFM synthetic and real data-set sequence \cite{Dense-NRSFM:CVPR-2013} introduced by Grag {\it{et al.}} Table \ref{tab:dense_3d_synthetic} lists the obtained 3D reconstruction error on these freely available standard synthetic data-sequence \cite{Dense-NRSFM:CVPR-2013}. Figure \ref{fig:Dense_result_synthetic} and \ref{fig:real_face_back_} provides visual insight of the procured 3D shapes over synthetic and real sequence respectively. To test the segmentation of different structure on real dataset, we performed experiments by combining two real dataset sequence. Figure \ref{fig:real_face_heart_} shows the segmentation of non-rigid feature tracks to their corresponding classes.  

\begin{table}
\begin{center}
\caption{Quantitative results on synthetic face sequence, without and with neighboring constraints.}
\label{tab:dense_3d_synthetic}
\begin{tabular}{|c|c|c|c|}
\hline
Dataset & No.of Feature points & e3d(C constraint) & e3d(CD constraint)\\ 
\hline
Face Sequence 1 & 3275 & 0.0749 & 0.0745\\
\hline
Face Sequence 2 & 3275 & 0.0506 & 0.050\\
\hline
Face Sequence 3 & 3275 & 0.0384 & 0.0380\\
\hline
Face Sequence 4 & 3275 & 0.0446 & 0.0443 \\
\hline
\end{tabular}
\end{center}
\end{table}

\section{Conclusions}
This paper has filled in a missing gap in the Structure-from-Motion family by proposing a new framework for ``Multi-body Non-rigid-Structure-from-Motion''. It achieves a joint non-rigid reconstruction and non-rigid shape segmentation of multiple deformable  structures observed in a single image sequence.  Under our new multi-body NRSFM framework, the solutions for motion segmentation can better constrain the solutions for 3D reconstruction.  This way, we achieved superior performance in both 3D non-rigid reconstruction and non-rigid motion segmentation, compared with the alternative, two stage method (first segment, then reconstruct).  In future, we plan to investigate the scalability issue observed in our current implementation; we aim to apply the new method to denser feature tracks in longer video sequences. In addition to this, we plan to improve over state-of-the-art 3D reconstruction method \cite{Dai-Li-He:CVPR-2012} to get much robust reconstruction for the objects undergoing non-rigid deformation.

\section*{Acknowledgments}
Y. Dai is funded in part by ARC Grants (DE140100180, LP100100588) and National Natural Science Foundation of China (61420106007). H. Li's research is funded in part by ARC grants (DP120103896, LP100100588, CE140100016) and NICTA (Data61).

\bibliographystyle{splncs}
\bibliography{Reference_Non_Rigid}

\end{document}